\documentclass[lettersize,journal]{IEEEtran}
\usepackage{amsmath,amsfonts}
\usepackage{algorithm}
\usepackage{algorithmicx}
\usepackage{algpseudocode}
\usepackage{array}
\usepackage[caption=false,font=normalsize,labelfont=sf,textfont=sf]{subfig}
\usepackage{textcomp}
\usepackage{stfloats}
\usepackage{url}
\usepackage{verbatim}
\usepackage{graphicx}
\usepackage{cite}
\usepackage{multirow}
\usepackage{amssymb}
\usepackage{bbm}
\usepackage{bibentry}
\usepackage{booktabs}
\usepackage{xcolor}
\usepackage{colortbl}
\usepackage{etoolbox}
\makeatletter
\newcommand*{\algrule}[1][\algorithmicindent]{%
  \makebox[#1][l]{%
    \hspace*{.2em}
    \vrule height .75\baselineskip depth .25\baselineskip
  }
}
\newcount\ALG@printindent@tempcnta
\def\ALG@printindent{%
    \ifnum \theALG@nested>0
        \ifx\ALG@text\ALG@x@notext
        \else
            \unskip
            \addvspace{-1pt}
            \ALG@printindent@tempcnta=1
            \loop
                \algrule[\csname ALG@ind@\the\ALG@printindent@tempcnta\endcsname]%
                \advance \ALG@printindent@tempcnta 1
            \ifnum \ALG@printindent@tempcnta<\numexpr\theALG@nested+1\relax
            \repeat
        \fi
    \fi
    }
\patchcmd{\ALG@doentity}{\noindent\hskip\ALG@tlm}{\ALG@printindent}{}{\errmessage{failed to patch}}
\makeatother
\DeclareMathOperator*{\argmax}{argmax} 

\begin{document}
\title{Diversify and Conquer: Open-set Disagreement for Robust Semi-supervised Learning with Outliers}

\author{Heejo Kong, Sung-Jin Kim, Gunho Jung, and Seong-Whan Lee$^*$, \IEEEmembership{Fellow, IEEE}
\thanks{Heejo Kong is with the Department of Brain and Cognitive Engineering, Korea University, Seoul, Republic of Korea (e-mail: hj\_kong@korea.ac.kr).

Sung-Jin Kim, Gunho Jung, and Seong-Whan Lee are with the Department of Artificial Intelligence, Korea University, Seoul, Republic of Korea (e-mail: s\_j\_kim@korea.ac.kr; gh\_jung@korea.ac.kr; sw.lee@korea.ac.kr).
}}
%

\markboth{Journal of \LaTeX\ Class Files,~Vol.~14, No.~8, August~2021}%
{Shell \MakeLowercase{\textit{et al.}}: A Sample Article Using IEEEtran.cls for IEEE Journals}


\maketitle

\begin{abstract}
Conventional semi-supervised learning (SSL) ideally assumes that labeled and unlabeled data share an identical class distribution, however in practice, this assumption is easily violated, as unlabeled data often includes unknown class data, \textit{i.e.}, outliers.
The outliers are treated as noise, considerably degrading the performance of SSL models.
To address this drawback, we propose a novel framework, Diversify and Conquer (DAC), to enhance SSL robustness in the context of open-set semi-supervised learning.
In particular, we note that existing open-set SSL methods rely on prediction discrepancies between inliers and outliers from a single model trained on labeled data.
This approach can be easily failed when the labeled data is insufficient, leading to performance degradation that is worse than naive SSL that do not account for outliers.
In contrast, our approach exploits prediction disagreements among multiple models that are differently biased towards the unlabeled distribution.
By leveraging the discrepancies arising from training on unlabeled data, our method enables robust outlier detection even when the labeled data is underspecified.
Our key contribution is constructing a collection of differently biased models through a single training process.
By encouraging divergent heads to be differently biased towards outliers while making consistent predictions for inliers, we exploit the disagreement among these heads as a measure to identify unknown concepts.
Extensive experiments demonstrate that our method significantly surpasses state-of-the-art OSSL methods across various protocols.
Codes are available at \url{https://github.com/heejokong/DivCon}.
\end{abstract}


\section{Introduction}
Semi-supervised learning (SSL) \cite{ssl_1,ssl_2,ssl_9} is a classical machine learning paradigm that leverages a large amount of unlabeled data alongside a limited set of labeled samples.
By utilizing readily available and inexpensive raw data, SSL reduces the dependency on extensive labeled datasets, which are time-consuming and labor-intensive for their construction \cite{data_5,data_6}.
However, the positive results in conventional SSL rely on the optimistic assumption that labeled and unlabeled data share an identical class distribution.
In practice, this assumption is often violated, as raw data typically includes out-of-class samples, known as outliers.
These outliers, treated as noise during subsequent training, directly corupt SSL models \cite{ssl_1,ossl_1}.
Thus, the models should not only classify samples from known categories, \textit{i.e.}, inliers, but also identify unknown samples as outliers.
This problem is known as open-set semi-supervised learning (OSSL) \cite{ossl_3}.
By identifying and excluding potential outliers from standard SSL processes, OSSL strives to avoid model corruption caused by noise accumulation.

\begin{figure}[t]
\begin{center}
\begin{tabular}{c@{}c@{}c}
\includegraphics[width=0.32\columnwidth]{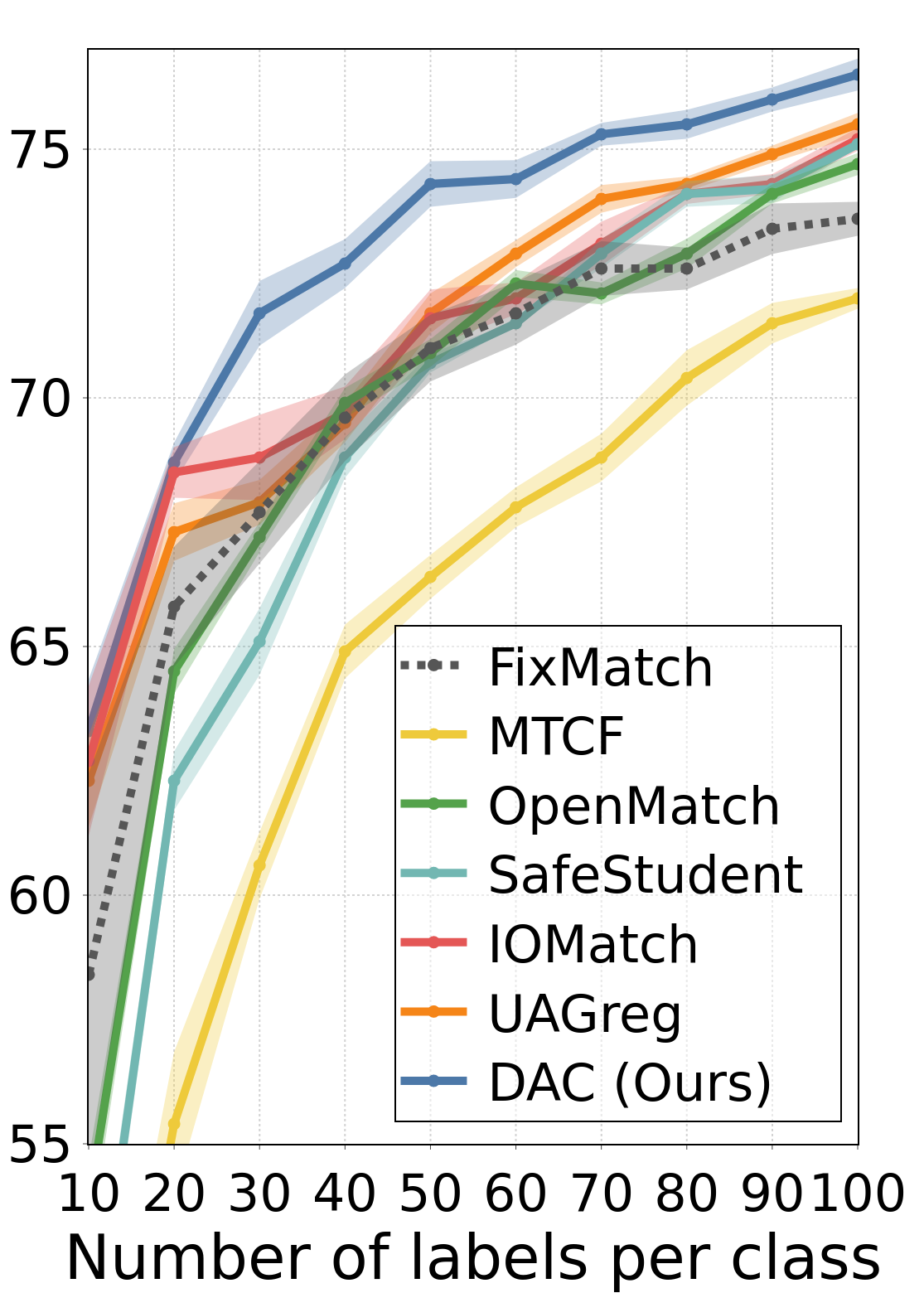} &
\includegraphics[width=0.32\columnwidth]{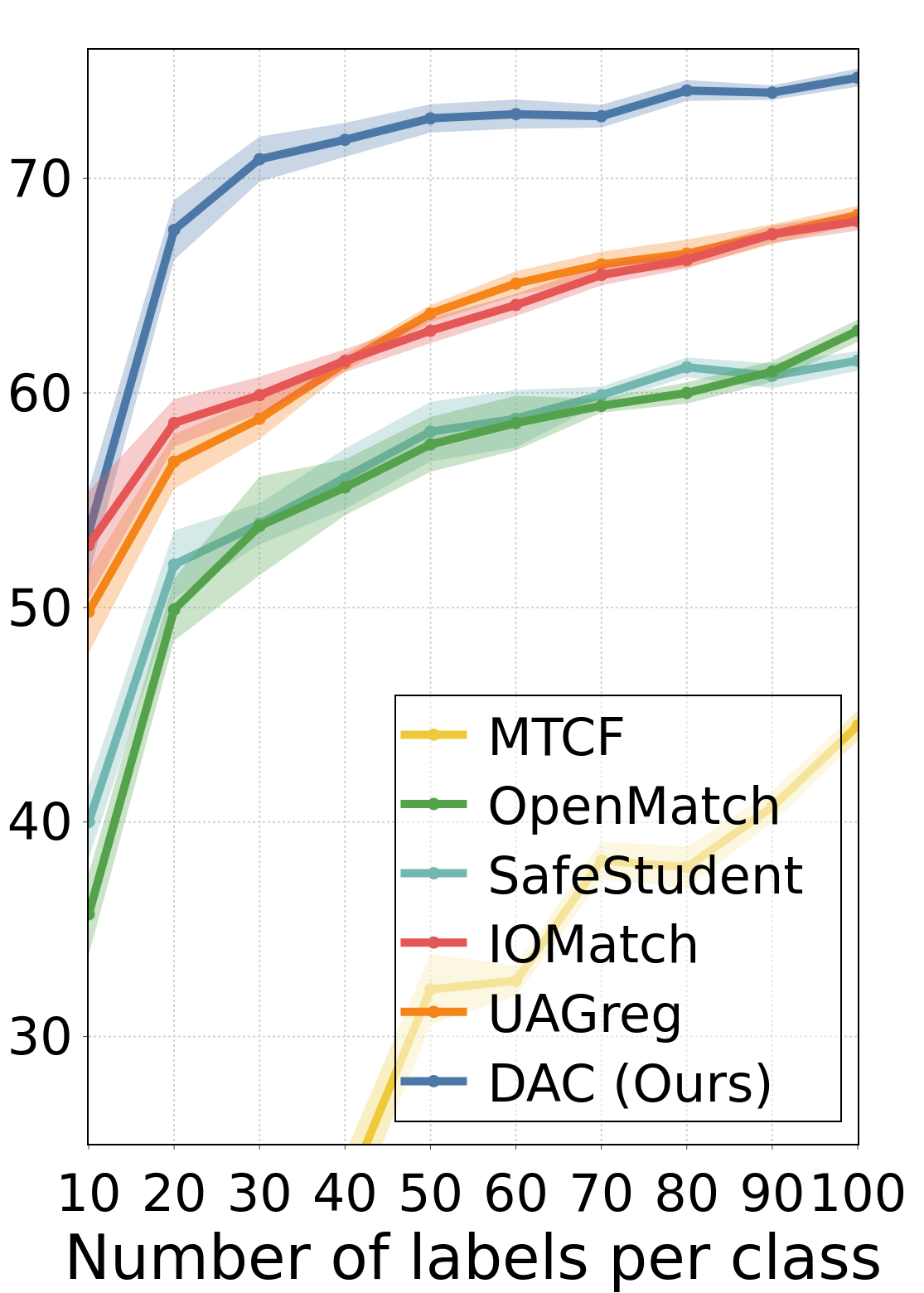} &
\includegraphics[width=0.32\columnwidth]{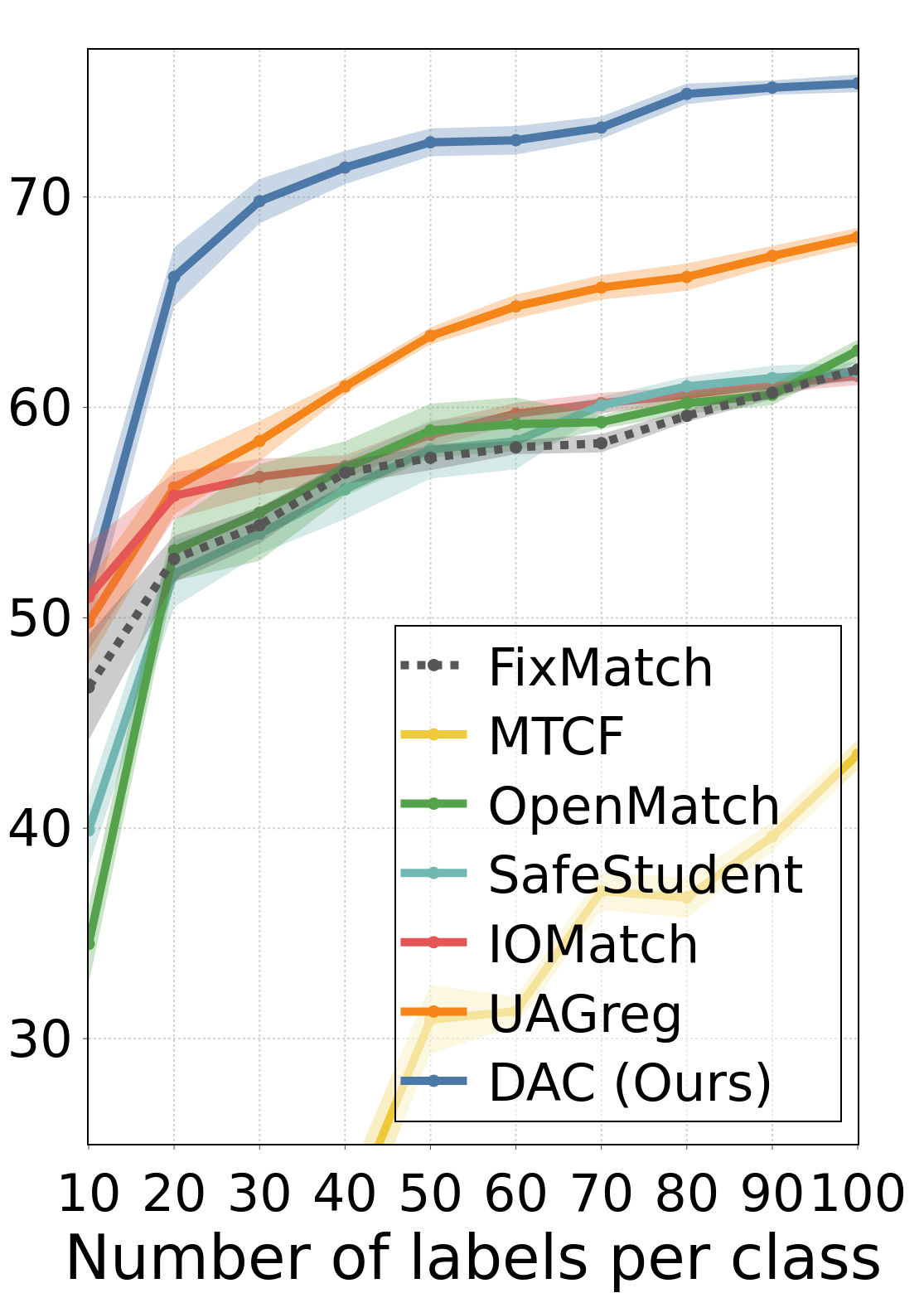} \\
(a) Closed-set acc. & (b) Open-set acc. & (c) Recall \\
\end{tabular}
\end{center}
\vspace{-0.2cm}
\caption{Comparison results for various open-set SSL algorithms (colored solid lines) trained on \textit{CIFAR-50} with varying number of labeled samples. As the number of available labeled samples decreases, the accuracy of existing algorithms significantly declines, and interestingly, some algorithms perform worse than a standard SSL baseline (black dotted lines) that does not account for outliers. Given this, we propose DAC, a novel paradigm that robustly addresses the open-set SSL problem even with sparse labeled knowledge.}
\vspace{-0.3cm}
\end{figure}

In OSSL, a key challenge lies in the absence of open-set knowledge; that is, no information is provided to distinguish unknown samples from known ones.
Recent works \cite{ossl_3,ossl_4,ossl_5,ossl_6,ossl_7,ossl_8} generally address this problem by leveraging prediction discrepancies from a single model trained with empirical risk minimization (ERM) on labeled data.
This approach is predicated on the hypothesis that, \textit{even for unseen data, inliers will achieve minimized risk through ERM, whereas outliers will not}.
Consequently, they consider samples with uncertain predictions on unlabeled data, which fail to minimize risk, as outliers.
By maximizing the expected probability for labeled data, while minimizing it for samples identified as outliers, they train the model to recognize the concept of the unknown.

However, the above hypothesis can be contradicted when labeled data is scarce.
More clearly, this work highlights underspecified cases where the ERM model struggles to generalize on unseen distributions due to the lack of sufficient labeled data.
In such cases, the model often fails to minimize the expected risk for inliers as well as outliers.
This leads the existing methods to predict most unlabeled data as outliers, resulting in an over-rejection problem that severely limits the utilization of valuable inliers in SSL.
As shown in Figure 1, existing open-set SSL methods, suffering from this over-rejection, significantly underperform than naive SSL methods that use all unlabeled data for training.
This underspecification is an inherent issue in SSL, which assumes limited labels, but it has rarely been addressed in the previous literature \cite{ossl_7}.

To address this drawback, we present a novel perspective, open-set disagreement, which emphasizes prediction inconsistencies among multiple models trained on unlabeled data.
This approach stems from experimental observations in conventional SSL.
As shown in Figure 2, SSL models trained on differently allocated unlabeled sets, despite sharing the same labeled data, produce divergent predictions for outliers.
This variability indicates that the attributes contributing to noise predictions, those assigning outliers to known categories, change with the differing outlier samples used for training.
That is, in contrast to inliers, which are anchored by labeled data, the noise accumulation associated with outliers can be manipulated.
From this observation, we derive the following hypothesis that, \textit{multiple SSL models, each biased differently towards the unlabeled data, will fail to make consistent predictions for outliers}.
This hypothesis forms the foundation of our study, using model disagreement as a signal for identifying potential outliers.
Our perspective capitalizes on the emergent differences between inliers and outliers during the training of unlabeled data, offering a robust solution to the OSSL challenge, even under conditions of underspecified labeled set.

\begin{figure*}[t]
\begin{center}
\begin{tabular}{c @{\quad} c @{\quad} c}
\includegraphics[height=3.5cm]{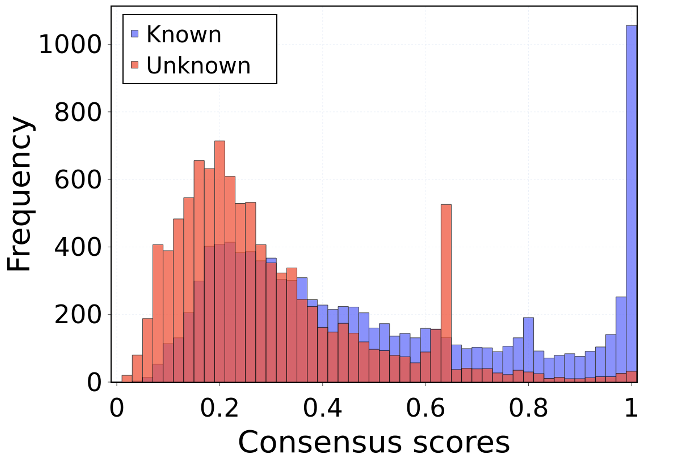} &
\includegraphics[height=3.5cm]{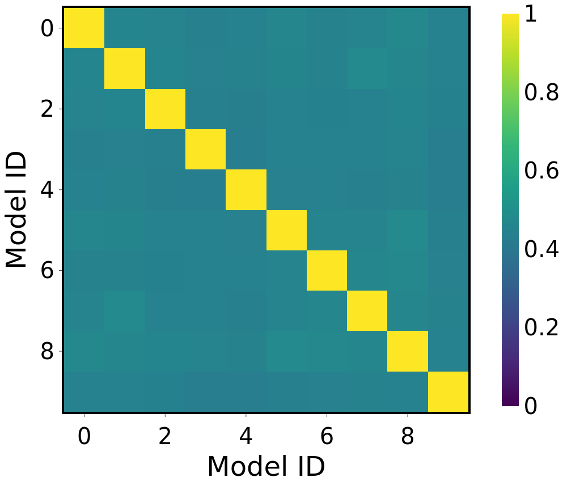} &
\includegraphics[height=3.5cm]{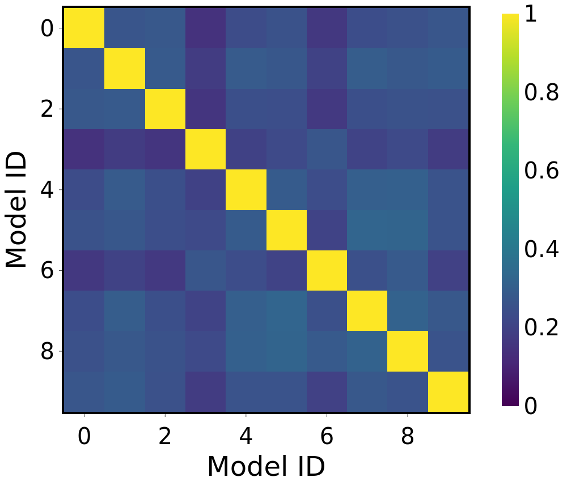} \\
\textcolor{black}{\small{(a) Histogram of consensus scores}} & \textcolor{black}{\small{(b) Prediction consensus for inliers}} & \textcolor{black}{\small{(c) Prediction consensus for outliers}} \\
\end{tabular}
\end{center}
\vspace{-0.2cm}
\caption{Qualitative results for multiple models trained with standard SSL baseline \cite{ssl_4} on the \textit{CIFAR-50-500}. Despite identical initialization parameters and labeled sample configurations, each model learns from a distinct subset of 20,000 unlabeled data samples; 50\% of these are from known classes, while the remainder are from unknown classes. (a) displays a histogram visualization of the normalized consensus scores calculated from Eq. (8). (b) and (c) demonstrate the prediction disagreements of multiple SSL models for inliers and outliers, respectively. The results suggest that multiple SSL models can be readily biased differently, particularly inducing substantial disagreements for outlier samples. Inspired by the observation, we introduce a novel perspective, open-set disagreement, which leverages the prediction consensus of diversely biased SSL models to identify potential outliers.}
\vspace{-0.2cm}
\end{figure*}
Building on the concept of open-set disagreement, we introduce a novel OSSL framework called Diversify and Conquer.
Instead of training multiple SSL models, which is inefficient due to high training costs, our approach constructs a collection of diverse functions through a single training process.
Our training strategy consists of three main components:
First, we adopt a multiple head structure, where all heads share the same feature extractor, and each head captures distinct functions.
Second, these heads are self-trained to predict known categories, similar to conventional SSL, while also minimizing mutual information to ensure diverse predictions on the unlabeled data.
This training encourages each head to focus on different attributes of the unlabeled data, resulting in diverse functions.
Third, to ensure training stability, the above objective is exploited only to update the divergent heads, while the feature extractor remains unchanged.
\textcolor{black}{Instead, the knowledge of open-set targets, estimated from the consensus scores of predictions, is distilled into the feature extractor.}
This enhances the prediction consistency of the heads for inliers while amplifying their distinctiveness for outliers.

Conclusively, we estimate sample-wise uncertainties by quantifying consensus among these predictions and identify potential outliers as samples with low consensus scores.
By suppressing their influence within the SSL objective, we prevent the model from being compromised.
Notably, the proposed DAC can be seamlessly integrated with existing SSL baselines without substantial loss in inference speed.
This demonstrates that our method can be an effective alternative in most cases when facing the existing open-set SSL problem.

Our main contributions can be summarized as follows:
\begin{itemize}
\item We introduce the novel perspective, open-set disagreement, leveraging prediction inconsistencies across multiple SSL models to identify outliers, enhancing the robustness of standard SSL approaches.
\item We propose the Diversify and Conquer framework, efficiently generating diverse models with only a single training process to estimate sample-wise uncertainties and detect potential outliers.
\item We validate our proposed method through extensive experiments, demonstrating that it outperforms state-of-the-art approaches, even when labeled data is underspecified.
\end{itemize}
\section{Motivation}
\noindent\textbf{Problem statement.}
This study addresses the open-set SSL problem, wherein both the labeled set $D_{l}$ and the unlabeled set $D_{u}$ are employed to train a classification model $f$.
Our aims is to train $f: \mathbb{R}^{H \times W \times 3} \rightarrow \mathbb{R}^{C}$ to predict the label $y\in \{1,\dots,C\}$ for a given input data $x$.
Unlike conventional SSL, open-set SSL operates under the assumption that the unlabeled set contains unknown class data, \textit{i.e.}, outliers, reflecting uncurated scenarios.
Consequently, the unlabeled set is partitioned into the inliers set $D_{u}^{\text{in}}$, which shares the class distribution with the labeled data, and the outliers set $D_{u}^{\text{out}}$, which does not.
In this context, our objective extends beyond classifying known categories to also identifying unknown classes data within the unlabeled set.
To this end, we integrate a detector $g: \mathbb{R}^{H \times W \times 3} \rightarrow \mathbb{R}$, tasked with predicting the binary property distinguishing inliers from outliers.

\subsection{Standard SSL with Outliers}
We first revisit conventional SSL frameworks that overlook outliers during training.
Given a batch of labeled samples $\mathcal{X} = \{(x_b, y_b)\}_{b=1}^{B}$ with one-hot labels $y_b$, and a batch of unlabeled samples $\mathcal{U} = \{u_b\}_{b=1}^{\mu B}$, where $\mu$ determines the relative size of $\mathcal{X}$ and $\mathcal{U}$, the frameworks jointly optimize the supervised loss $L_{s}$ and the unsupervised loss $L_{u}$ as follows:
\begin{equation}
L_{\text{ssl}}(f)=L_s(f) + \lambda_u L_u(f),
\end{equation}
where $\lambda_{u}$ is a hyperparameter to control the weight of the unsupervised objective.
$L_{s}$ is defined as the cross-entropy loss $H(\cdot, \cdot)$ between ground-truth labels and model’s predictions:
\begin{equation}
L_s(f)=\frac{1}{B} \sum_{b=1}^B H(y_b, p(f|A_w(x_b))),
\end{equation}
where $A_{w}(\cdot)$ refers to a random augmentation function to obtain the weakly augmented view.
Following previous consistency frameworks \cite{ssl_3,ssl_4,ssl_14}, we define $L_{u}$ as the cross-entropy between pseudo-labels and model’s predictions:
\begin{equation}
L_u(f)=\frac{1}{\mu B} \sum_{b=1}^{\mu B} \mathbbm{1}(\max p_{b} \geq \tau) \cdot H(\hat{y}_b, p(f|A_s(u_b))),
\end{equation}
where $A_{s}(\cdot)$ denotes a strong augmentation function and $p_b=p(f|A_w(u_b))$ represents the probability estimated from the weakly augmented view of $u_b$.
\textcolor{black}{$\hat{y}_b=\argmax{p_b}$ indicates the predicted one-hot pseudo-label.}
Here, $\tau$ is a pre-defined threshold ensuring that only pseudo-labels with high confidence are retained.
This approach optimizes the model by minimizing entropy solely for those unlabeled data points with confident predictions during the current training step.

\begin{figure*}[t]
    \centering
\includegraphics[width=0.90\linewidth]{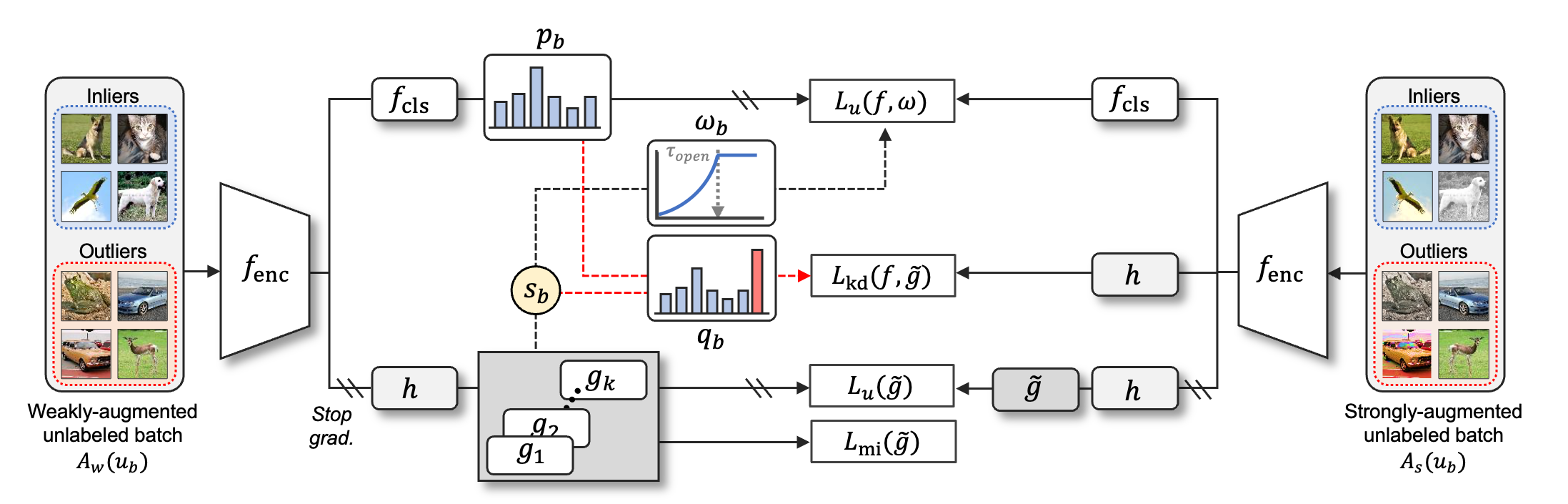}
\vspace{-0.2cm}
\caption{An overview of the proposed framework, DAC, for the unlabeled data training. Our key components, contrasting with conventional SSL, include the consensus scores $s_b$, which estimates the likelihood of a sample being an outlier, and  divergent heads $\tilde{g}$ that predict the scores. By leveraging $s_b$, we employ a reweighting strategy (Eq. (13)) that diminishes the impact of outliers in the unsupervised objective, thus safeguarding the target classifier from being compromised by outliers. Note that our diversification strategy $L_{\text{mi}}$ (Eq. (5)) potentially disrupts the training stability of the feature extractor $f_{\text{enc}}$. Consequently, we employ this strategy specifically for training $\tilde{g}$, while introducing $L_{\text{kd}}$ (Eq. (12)), which directly propagates the estimated open-set knowledge to $f_{\text{enc}}$.}
\vspace{-0.2cm}
\end{figure*}
However, deep models frequently exhibit overconfidence in their predictions, even when they are incorrect \cite{tnnls_1}.
Consequently, existing SSL methods are prone to confirmation bias \cite{cb_1,ssl_4}, where the model reinforces initial erroneous predictions despite their inaccuracy.
This bias also affects outliers: if outliers are initially misclassified with high confidence as known classes, they are treated as noise and consistently mispredicted in subsequent training.
This leads to persistent incorrect predictions for outliers with similar attributes.

\subsection{Deriving Open-set Disagreement}
This study explores the differing confirmation biases towards outliers compared to inliers.
As shown in Figure 2, models trained with standard SSL produce divergent predictions for outliers as the unlabeled data varies, despite identical initialized parameters and given labeled samples.
This suggests that models develop distinct biases based on various attributes, generating noisy predictions for outliers.
In contrast, predictions for inliers, guided by labeled data, remain consistent across models.
Based on this observation, we hypothesize that conventional SSL can be adapted to construct multiple models with varying biases towards unlabeled data, and these models fail to consistently predict outliers.
This hypothesis supports the introduction of open-set disagreement, a novel perspective for identifying unknown concepts within the open-set SSL paradigm.
If we can construct a collection of models with different biases towards unlabeled data, the prediction consistency of these models can serve as a measure for detecting outliers.
That is, a sample receiving low consensus predictions from multiple models is likely an unknown sample.

We assert that the proposed perspective can serve as a robust outlier detection strategy, even when labeled data is underspecified.
Most existing open-set SSL methods detect differences between inliers and outliers during the training on labeled data and transfer this knowledge to the target classifier.
However, this approach is effective only when the labeled data can highlight differences concerning unknown concepts.
When labeled knowledge is sparse, this often leads to over-rejection problem, where both inliers and outliers fail to generalize, resulting in most samples being considered outliers.
In contrast, our strategy focuses on the differences between inliers and outliers manifested during training on unlabeled data.
Although conventional SSL may induce overconfidence towards outliers, it still effectively minimizes the expected risk for inliers.
Open-set disagreement operates as a voting ensemble, enabling uncertainty modeling for samples even if individual models are overconfident about outliers.
This strategy is more robust to failures in learning from inliers and can be a viable alternative when labeled data is underspecified.
\begin{figure}[t]
    \centering
\includegraphics[width=0.85\columnwidth]{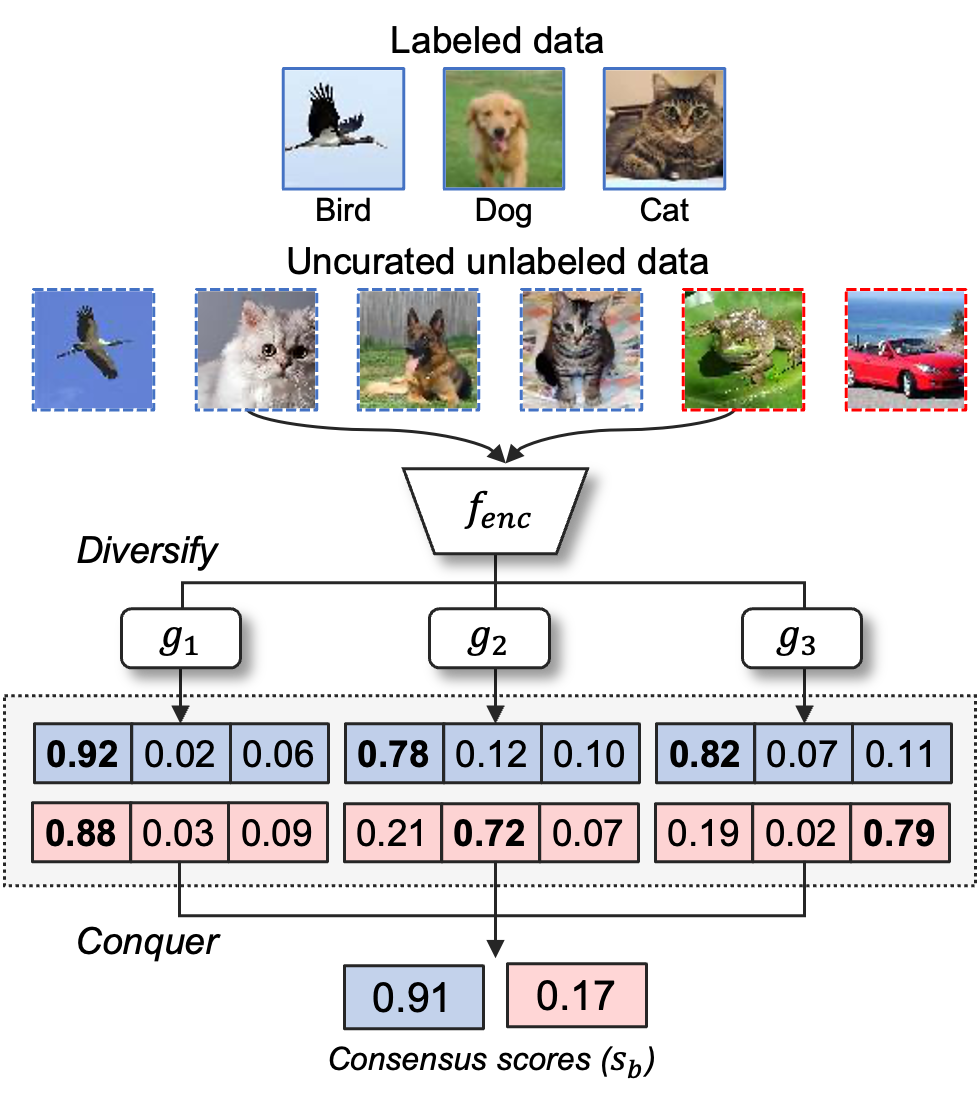}
\vspace{-0.3cm}
\caption{A conceptual illustration of estimating consensus scores with divergent heads. If the predictions of the heads for a specific sample are consistent, \textit{i.e.}, the consensus score is close to 1, the data is considered an inlier. Conversely, if the consensus score is low, the sample is considered an outlier.}
\vspace{-0.3cm}
\end{figure}
\section{Proposed Method}
\subsection{Approach Overview}
In this section, we present our open-set SSL framework.
Our model consists of a shared encoder $f_{\text{enc}}(\cdot)$ with two components: (1) the target head $f_{\text{cls}}(\cdot)$ for closed-set classification and (2) divergent heads $\tilde{g}(\cdot)$ to estimate the uncertainty of each prediction for open-set recognition.
Hence, the classifier $f=f_\text{cls}(f_\text{enc}(\cdot))$ can be represented as the composition of a feature extractor $f_{\text{enc}}$ and a target head $f_{\text{cls}}$.

An overview of our proposed framework is shown in Figure 3.
For the SSL objective, we employ FixMatch \cite{ssl_4}, a simple yet effective approach aligned with Eq. 1.
Unlike conventional SSL, this work addresses the open-set SSL setting, where the unlabeled dataset includes out-of-class data, \textit{i.e.}, outliers.
To mitigate the risk of model corruption caused by the outliers, we reformulate the unsupervised loss in Eq. 3 using uncertainty scores derived from the divergent heads:
\begin{equation}
L_u(f,\omega)=\frac{1}{\mu B} \sum_{b=1}^{\mu B} \omega(u_b) \cdot \mathbbm{1}(\max p_{b} \geq \tau) \cdot H(\hat{y}_b, p(f|A_s(u_b))),
\end{equation}
where $\omega(\cdot) \in (0,1)$ denotes a weighting function that reduces the influence of potential outliers by assigning them values near $0$.
Hence, the novelty of our approach lies in the sophisticated design of this weight function using divergent heads and a tailored training strategy.

\subsection{Diversify and Conquer}
To estimate the uncertainty of a specific datapoint belonging to the out-of-class distribution, we utilize the perspective of open-set disagreement.
Hence, we consider samples with inconsistent predictions from multiple functions, each biased differently towards the unlabeled distribution, as outliers.
The core of our idea comprises two strategies:
\textit{i)} Diversify: a method to construct diverse functions by encouraging disagreements.
\textit{ii)} Conquer: an approach to estimate uncertainty by quantifying the consistency of the functions' predictions.

\noindent\textbf{Diversify: Train divergent heads.}
\textcolor{black}{
The simplest way to construct diverse functions is to independently apply SSL to each model, which demands significant computational resources, requiring separate training and inference for all models.
For a set of $K$ models, this approach incurs at least $K$-fold computational cost compared to conventional SSL methods.
To address this inefficiency, we adopt a multi-head structure $\tilde{g}=\{g_i|i=1,\dots,K\}$ with a shared encoder, allowing diverse functions to be constructed within a single training process.
To achieve this, we define a diversification objective $L_{\text{div}}$ that ensures each head develops distinct biases towards the unlabeled distribution.
}
\textcolor{black}{
The proposed objective combines the conventional SSL loss $L_{\text{ssl}}$ with a mutual information loss $L_{\text{mi}}$, balanced by the hyperparameter $\lambda_{\text{mi}}$ as:
\begin{equation}
L_{\text{div}}(\tilde{g})=\sum_i L_{\text{ssl}}(g_i)+\lambda_{\text{mi}} \sum_{i \neq j} L_{\text{mi}}(g_i, g_j).
\end{equation}
The preceding loss, $L_{\text{ssl}}$, induces each head to be biased with respect to the unlabeled distribution, and the subsequent loss, $L_{\text{mi}}$, induces their biases to be distinct from each other.
}

\textcolor{black}{
Specifically, each head $g_{i}: \mathbb{R}^{d^{\prime}} \rightarrow \mathbb{R}^{C}$ predicts probabilities for known classes using embeddings $z_b=h(f_{\text{enc}}(A_{\text{w}}(u_b)))$, where the shared encoder $f_{\text{enc}}: \mathbb{R}^{H \times W \times 3} \rightarrow \mathbb{R}^{d}$ extracts features, and the non-linear projection head $h: \mathbb{R}^{d} \rightarrow \mathbb{R}^{d^{\prime}}$ maps them to low-dimensional representations.
By minimizing the expected risk on labeled data while introducing self-training on unlabeled distributions, we allow each head to induce a confirmation bias.
The SSL objective consists of a combination of supervised and unsupervised losses, similar to Eq. 1: $L_{\text{ssl}}(g_i)=L_s(g_i) + \lambda_u L_u(g_i)$.
}

\textcolor{black}{
In the motivation experiment, we explored obtaining diverse functions by training each model on distinct subsets of the unlabeled data.
However, since our approach utilizes a shared encoder, simply assigning different subsets to each head is insufficient to ensure diverse biases in their predictions.
Moreover, the previous approach heavily relies on the specific composition of the assigned subsets, making it challenging to achieve consistent and robust training across the functions.
Hence, rather than relying on different subset assignments, we train each pair of heads to generate statistically independent predictions by minimizing their mutual information.
By comparing predictions on unlabeled data while optimizing the SSL objective, each function is encouraged to leverage distinct predictive characteristics of the unlabeled distribution, thereby inducing diverse biases:
\begin{equation}
L_{\text{mi}}
(g_i,g_j)=\mathbb{E}_{z_b \sim \mathcal{Z}}[D_{\text{KL}}(p(g_i, g_j|z_b) \| p(g_i|z_b) \otimes p(g_j|z_b))],
\end{equation}
where $D_{KL}(\cdot||\cdot)$ denotes the KL divergence, and $z_b \sim \mathcal{Z}$ represents the set of extracted embeddings.
We optimize this quantity using empirical estimates of the joint distribution $p(g_i,g_j|z_b)$ and the product of marginal distributions $p(g_i|z_b) \otimes p(g_i|z_b)$, following previous works \cite{mi_1,mi_2,mi_3} adopting the mutual information for their objectives.
}

\noindent\textbf{Conquer: Measure consensus scores.}
Then, by leveraging the diverse set of functions, we measure uncertainty scores for outlier detection.
The core of open-set disagreement is that diversifying predictions will lead to functions that disagree on open-set data.
Thus, we consider samples with high disagreement on $\tilde{g}$ as likely outliers, indicating high uncertainty.
A conceptual illustration of this strategy is depicted in Figure 4.

Specifically, we quantify the consensus scores $\hat{s}_{b} \in \mathbb{R}$ with the pairs of estimated predictions by the model collection $\tilde{g}$.
The score is derived as the expectation of the exponential negative L1 distances between all pairs of divergent heads as:
\begin{equation}
\hat{s}_b=\mathbb{E}_{(g_i, g_j) \sim \tilde{g}}\left[\exp\left(-||p(g_i|z_b)-p(g_j|z_b)||\right)\right].
\end{equation}
Intuitively, the higher scores are assigned as the predictions estimated by the heads for a specific datapoint $u_{b}$ be more similar.
Hence, a lower score indicates higher sample-wise uncertainty.
\textcolor{black}{
To utilize the estimated uncertainty scores as weight functions in Eq. 4, we perform two post-processing on them.
The first process is min-max normalization, which allows us to directly leverage the scores as probabilities of belonging to a known distribution.
To this end, the scores are stored in a memory queue $\mathcal{S} = \{s_{n}\}_{n=1}^{N}$, where $N$ is the total number of unlabeled samples.
Each score is derived by a linear transformation based on the minimum $s_{\text{min}}=\min \mathcal{S}$ and maximum $s_{\text{max}}=\max \mathcal{S}$ values of the all scores as:
\begin{equation}
\bar{s}_{b}=\left(\hat{s}_b-s_{\text{min}}\right) /\left(s_{\text{max}}-s_{\text{min}}\right).
\end{equation}
}
\textcolor{black}{
As a follow-up process, we adopt an exponentially moving average (EMA) uncertainty, which acts as a warm-up process to ensure the stable utilization of the scores in the early stage of training.
All initial values $s_{b}$ are set to $0$, and they are updated incrementally with the momentum parameter $\alpha$ as:
\begin{equation}
s_{b} \leftarrow \alpha \cdot s_{b}+(1-\alpha) \cdot \bar{s}_{b}.
\end{equation}
}

\noindent\textbf{Open-set knowledge Distillation.}
\textcolor{black}{
This work leverages the fact that differently biased multiple functions exhibit relatively varying degrees of prediction consistency for inliers and outliers.
Hence, the proposed objective $L_{\text{div}}$ explicitly encourages each head to produce distinct predictions on the unlabeled distribution.
However, this simultaneously hinders the model from converging to consistent predictions for the known classes contained within the unlabeled data.
Since our architecture shares a common feature extractor $f_{\text{enc}}$ among all heads, this strategy may degrade the performance of the target classifier $f_{\text{cls}}$.
To mitigate this risk, we use the gradient derived from $L_{\text{div}}$ only to update the disagreeing heads $\tilde{g}$.
Instead, we introduce an additional strategy that distills the estimated open-set knowledge into the shared feature extractor.
By exploiting the outliers as a single generic class, we attempt to train a more sophisticated feature extractor.
}

Specifically, we exploit the derived score $s_{b}$ to generate open-set targets that integrate the expected probabilities of the target classifier with a binary indicator for out-of-class samples.
Let $p_{b,c}$ denotes the probability for the $c$-th known class predicted by the target head $f_{\text{cls}}$.
From the perspective of open-set disagreement, the score is close to 0 for samples unlikely belonging to any known classes.
Hence, our framework consider this score as the negative probability for $(C+1)$-th class by treating all unknown samples as an one virtual class.
The probabilities are combined into a $(C+1)$-way class probability distribution $q_b \in \mathbb{R}^{C+1}$ as follows:
\begin{equation}
q_{b, c}= \begin{cases}
p_{b, c} \cdot s_{b} & \text { if } 1 \leq c \leq C \\ (1-s_{b}) & \text { if } c=C+1
\end{cases}.
\end{equation}

Following \cite{ssl_5,ssl_7,ssl_17}, we propagate instance-wise similarity relationships to the feature extractor, akin to graph-based SSL approaches.
OSSL does not provide labels for unknown concepts, making predictions for single outlier samples generally unreliable.
Thus, we stabilize the distillation of open-set knowledge by forming an adjacency graph based on embedding similarities and promoting smooth predictions that reflect this graph.
In our implementation, the embedding $z_b$ and open-set target $q_b$ are stored in memory banks $Q_{z} \in \mathbb{R}^{d^{\prime} \times M}$ and $Q_{b} \in \mathbb{R}^{(C+1) \times M}$, respectively.
The memory bank is updated with a first-in-first-out strategy and contains only the extracted outputs for unlabeled samples $u_b$.
The affinity of embeddings between the current sample and the $i$-th sample in the memory bank is measured as:
\begin{equation}
e_{b, i}=\frac{\exp (\bar{z}_b^{\prime} \cdot \bar{z}_i / t_{e})}{\sum_{j=1}^M \exp (\bar{z}_b^{\prime} \cdot \bar{z}_j / t_{e})},
\end{equation}
where $M$ denotes the size of the memory bank, and $t_{e}$ is the scalar temperature parameter for sharpening.
To calculate the affinity similarity, we utilize normalized embeddings $\bar{z}_{b}$.
The embedding $\bar{z}_b^{\prime}$ is obtained from the strongly augmented view for the current sample.
Finally, the objective is derived as the consistency loss between the predicted targets $q_b$ and the aggregated label $\sum_{i=1}^{M} q_i \cdot e_{b,i}$ as follow:
\begin{equation}
L_{\text{kd}}(f,\tilde{g})=-\frac{1}{\mu B} \sum_{b=1}^{\mu B}q_b \cdot \log (\sum_{i=1}^M q_i \cdot e_{b, i}).
\end{equation}

\textcolor{black}{
This training procedure encourages the feature extractor to push the features of samples, on which multiple heads disagree, away from the clusters of known categories in the embedding space.
By encouraging the samples predicted similarly by divergent heads to converge to the similar point in the embedding space, we effectively enhance the distinction between inliers and outliers.
}

\subsection{Semi-supervised Learning with DAC}
Ultimately, by utilizing the uncertainty scores estimated by the proposed DAC, we aim to prevent model degradation caused by potential outliers.
To achieve this, we introduce a reweighting strategy that reduces the influence of samples likely to be outliers in the unsupervised objective.
Unlike existing OSSL methods, our approach employs a soft rejection that differentially excludes outliers rather than completely excluding them.
We observed that excessive rejection of outliers is more detrimental to SSL performance.
The proposed strategy effectively addresses the over-rejection problem by enhancing the recall of ambiguous but valuable inliers, which are difficult to distinguish from unknown concepts.

Let the set of estimated scores for the entire unlabeled data be $\mathcal{S} = \{s_{n}\}_{n=1}^{N}$.
To automatically determine an open-set threshold $\tau_{\text{open}}$ for distinguishing potential outliers, we exploit Otsu’s thresholding \cite{otsu}.
$\tau_{\text{open}}$ is designed as the threshold value that optimizes binary classification on the set $\mathcal{S}$.
We consider samples with the estimated scores higher than $\tau_{\text{open}}$ as inliers and assign a weighting function of 1 to these samples.
The other samples are considered outliers, and their weights are adjusted progressively according to their assigned scores.
Hence, the proposed reweighting function is derived as:
\begin{equation}
\omega(u_b)=
\begin{cases}
1 & \text{if } s_{b} \geq \tau_{\text{open}} \\
{(s_{b} / \tau_{\text{open}})}^{t_{w}} & \text{otherwise}
\end{cases},
\end{equation}
where $t_{w}$ represents a hyperparameter used to adjust the strength of the soft rejection.
Note that as the temperature scale approaches 0, it represents standard SSL without rejecting outliers, whereas higher temperatures indicate hard rejection.

\noindent\textbf{Overall objectives.}
The overall training objective of our proposed OSSL framework is the weighted sum of the supervised loss $L_s$, the unsupervised loss $L_u$ with weighting function $\omega$, the diversifying loss $L_{\text{div}}$, and the open-set knowledge distillation loss $L_{\text{kd}}$ with hyperparameters $\lambda_{u}$, $\lambda_{\text{mi}}$, and $\lambda_{\text{kd}}$.
Hence, the overall loss can be written as follows:
\begin{equation}
L_{\text{total}}=L_s(f) + \lambda_u L_u(f,\omega) + L_{\text{div}}(\tilde{g}) + \lambda_{\text{kd}} L_{\text{kd}}(f,\tilde{g}).
\end{equation}
\begin{table*}[t]
{
\normalsize
\centering

\begin{tabular}{p{1.0\textwidth}}
\toprule
\textbf{Algorithm 1} Optimization procedure of the proposed framework \\ \midrule
\textbf{Input:} Set of labeled data $D_l$ and unlabeled data $D_u$. Shared encoder $f_{\text{enc}}(\cdot)$. Target classifier $f_{\text{cls}}(\cdot)$. Divergent heads $\tilde{g}(\cdot)$. \\
\hspace*{\algorithmicindent}\quad\, Non-linear projection head $h(\cdot)$. Learning rate $\eta$. EMA momentum $\alpha$. Embedding and label buffers $Q_z$ and $Q_q$.
\begin{algorithmic}[1]
\State {\textbf{Initialize}} a set of model parameters $\theta$, and a consensus memory queue $\mathcal{S}=\{s_n=0|n=1,...N\}$
\While{\textit{model not converge}}
\For{$iter=1$ to $I_{\text{max}}$}
    \State {\textbf{Sample}} a batch of labeled data $\mathcal{X}=\{(x_b,y_b)\}_{b=1}^{B}$ and unlabeled data $\mathcal{U}=\{(u_b)\}_{b=1}^{\mu B}$
    \State $L_{\text{total}}=L_{s}(f)+L_{\text{div}}(\tilde{g})$ \Comment{Eq. 2 and 5}
    \State {\textbf{Update}} $s_b \leftarrow \alpha \cdot s_b + (1-\alpha) \cdot \bar{s}_b$ 
    for an unlabeled batch $\mathcal{U}$ \Comment{Eq. 9}
    \State {\textbf{Compute}} unified open-set targets $q_b$ with the updated scores $s_b$ \Comment{Eq. 10}
    \If {$iter > I_{\text{warm\_up}}$}
        \State {\textbf{Compute}} sample-wise uncertainty scores $\omega(u_b)$ \Comment{Eq. 13}
        \State $L_{\text{total}}=L_{\text{total}}+\lambda_{u}L_u(f,\omega)+\lambda_{\text{kd}}L_{\text{kd}}(f,\tilde{g})$ \Comment{Eq. 4 and 12}
    \EndIf
    \State {\textbf{Update}} $\theta \leftarrow \theta - \eta \nabla_{\theta} L_{\text{total}}$
    \State {\textbf{Update}} $Q_z$ with embeddings $\bar{z}_b$ and $Q_q$ with predicted labels $q_b$
\EndFor
\EndWhile
\end{algorithmic}
\textbf{Output:}
Trained model parameters $\theta$  \\ \bottomrule
\end{tabular}

}
\vspace{-0.3cm}
\end{table*}
\section{Experiments}
\subsection{Experimental Settings}
\noindent\textbf{Setup.}
We evaluate the effectiveness of our proposed method on several SSL image classification benchmarks, similar to the previous open-set SSL works \cite{ossl_4,ossl_7,ossl_8}.
Specifically, we conduct extensive experiments with varying amounts of labeled data and varying number of class splits (known/unknown classes) on public datasets, CIFAR-10/100 \cite{data_1}, and ImageNet \cite{data_2}.
All experiments are repeated three times with different initial seeds, and the results are reported the average performance with standard deviation.
Each experiment is conducted on four NVIDIA A40 GPUs, each with 48GB of memory.

\noindent\textbf{Compared baselines.}
We compared the proposed DAC with various representative SSL and OSSL methods.
For standard SSL baselines, we consider the latest state-of-the-arts, including MixMatch \cite{ssl_3}, FixMatch \cite{ssl_4}, CoMatch \cite{ssl_5}, SimMatch \cite{ssl_6}, and SoftMatch \cite{ssl_7}.
\textcolor{black}{For open-set SSL baselines, we employ the published works, including MTCF \cite{ossl_3}, OpenMatch \cite{ossl_4}, Safe-Student \cite{ossl_5}, OSP \cite{ossl_10}, SSB \cite{ossl_11}, IOMatch \cite{ossl_7}, and UAGreg \cite{ossl_8}.}
We exclude earlier OSSL methods, UASD \cite{ossl_2} and D3SL \cite{ossl_1}, since their results perform worse than a model trained only with labeled data on open-set SSL settings.

To ensure the rigorous fairness in evaluation comparisons, our implementations adhere to the unified test bed from prior work \cite{ossl_7}, utilizing the USB codebase \cite{ssl_18}. 
For standard SSL methods, we employed re-implementations from the codebase, as they outperform the originally published results under standard SSL conditions.
Open-set SSL methods were incorporated into our test bed using their released code.
Hyperparameters are set to the optimal values reported in the original papers.
All experiments were conducted with the same backbone, data splits, and random seeds across each setting.

\newcolumntype{b}{>{\color{black}}c}
\begin{table*}[t]
\caption{Closed-set classification accuracy (\%) on known-class test data across different class splits and labeled set sizes. \\Results are reported as the average with standard deviation over 3 runs on different random seeds.}
\begin{center}
\resizebox{1.0\textwidth}{!}
{
\begin{tabular}{lbccbccbcccc}
\toprule
Dataset & \multicolumn{3}{c}{CIFAR-10} & \multicolumn{6}{c}{CIFAR-100} & \multicolumn{2}{c}{ImageNet-30}  \\
\cmidrule(lr){1-1}\cmidrule(lr){2-4}\cmidrule(lr){5-10}\cmidrule(lr){11-12}
\# of konwn / unknown & \multicolumn{3}{c}{6 / 4} & \multicolumn{3}{c}{20 / 80} & \multicolumn{3}{c}{50 / 50} & \multicolumn{2}{c}{20 / 10} \\
\cmidrule(lr){1-1}\cmidrule(lr){2-4}\cmidrule(lr){5-7}\cmidrule(lr){8-10}\cmidrule(lr){11-12}
\# of labeled samples & 5 & 10 & 25 & 5 & 10 & 25 & 5 & 10 & 25 & 1\% & 5\% \\ \midrule
Labeled only & 26.8\scriptsize{$\pm$4.9} & 30.4\scriptsize{$\pm$5.3} & 34.5\scriptsize{$\pm$3.1} & 26.6\scriptsize{$\pm$4.0} & 30.3\scriptsize{$\pm$4.2} & 44.9\scriptsize{$\pm$3.8} & 23.5\scriptsize{$\pm$5.1} & 35.2\scriptsize{$\pm$5.2} & 45.1\scriptsize{$\pm$4.0} & 30.1\scriptsize{$\pm$2.2} & 55.8\scriptsize{$\pm$1.3} \\ \midrule
MixMatch \cite{ssl_3} & 34.7\scriptsize{$\pm$3.3} & 36.4\scriptsize{$\pm$4.1} & 48.7\scriptsize{$\pm$2.9} & 31.1\scriptsize{$\pm$4.9} & 41.4\scriptsize{$\pm$3.7} & 52.8\scriptsize{$\pm$2.1} & 30.3\scriptsize{$\pm$1.4} & 36.3\scriptsize{$\pm$2.7} & 52.8\scriptsize{$\pm$1.9} & 32.8\scriptsize{$\pm$2.3} & 62.2\scriptsize{$\pm$2.0} \\
FixMatch \cite{ssl_4} & 85.6\scriptsize{$\pm$5.6} & 88.5\scriptsize{$\pm$2.3} & 92.4\scriptsize{$\pm$0.8} & 46.9\scriptsize{$\pm$2.1} & 54.0\scriptsize{$\pm$2.5} & 66.1\scriptsize{$\pm$0.9} & 50.4\scriptsize{$\pm$1.5} & 59.8\scriptsize{$\pm$1.4} & 69.9\scriptsize{$\pm$0.9} & 72.7\scriptsize{$\pm$3.6} & 87.7\scriptsize{$\pm$1.5} \\
CoMatch \cite{ssl_5} & 86.9\scriptsize{$\pm$1.8} & 87.7\scriptsize{$\pm$2.5} & 90.3\scriptsize{$\pm$1.4} & 51.5\scriptsize{$\pm$3.1} & 59.7\scriptsize{$\pm$1.8} & 66.9\scriptsize{$\pm$0.7} & 47.8\scriptsize{$\pm$1.7} & 58.8\scriptsize{$\pm$1.8} & 69.4\scriptsize{$\pm$1.2} & 82.7\scriptsize{$\pm$0.9} & 91.0\scriptsize{$\pm$0.5} \\
SimMatch \cite{ssl_6} & 85.1\scriptsize{$\pm$4.1} & 88.2\scriptsize{$\pm$2.3} & 90.1\scriptsize{$\pm$1.7} & 47.9\scriptsize{$\pm$3.8} & 56.9\scriptsize{$\pm$2.1} & 63.8\scriptsize{$\pm$2.3} & 54.2\scriptsize{$\pm$2.1} & 63.8\scriptsize{$\pm$0.9} & 69.8\scriptsize{$\pm$1.3} & 82.7\scriptsize{$\pm$1.1} & 91.3\scriptsize{$\pm$0.3} \\
SoftMatch \cite{ssl_8} & 77.6\scriptsize{$\pm$3.1} & 85.2\scriptsize{$\pm$3.2} & 90.5\scriptsize{$\pm$1.1} & 52.3\scriptsize{$\pm$1.9} & 59.3\scriptsize{$\pm$2.1} & 65.7\scriptsize{$\pm$1.2} & 54.1\scriptsize{$\pm$1.1} & 63.1\scriptsize{$\pm$1.5} & 68.6\scriptsize{$\pm$0.9} & 78.4\scriptsize{$\pm$2.1} & 90.2\scriptsize{$\pm$0.7} \\
\midrule
MTCF \cite{ossl_3} & 42.3\scriptsize{$\pm$4.2} & 47.9\scriptsize{$\pm$4.1} & 67.5\scriptsize{$\pm$2.1} & 41.8\scriptsize{$\pm$4.4} & 43.2\scriptsize{$\pm$3.4} & 53.4\scriptsize{$\pm$1.7} & 42.8\scriptsize{$\pm$2.9} & 46.6\scriptsize{$\pm$1.1} & 56.9\scriptsize{$\pm$0.5} & 48.2\scriptsize{$\pm$3.8} & 81.6\scriptsize{$\pm$1.3} \\
OpenMatch \cite{ossl_4} & 47.3\scriptsize{$\pm$3.3} & 53.7\scriptsize{$\pm$2.3} & 61.8\scriptsize{$\pm$1.9} & 45.9\scriptsize{$\pm$3.1} & 56.1\scriptsize{$\pm$2.1} & 66.9\scriptsize{$\pm$1.8} & 52.4\scriptsize{$\pm$1.1} & 54.1\scriptsize{$\pm$0.9} & 68.0\scriptsize{$\pm$0.5} & 55.8\scriptsize{$\pm$3.5} & 63.5\scriptsize{$\pm$1.0} \\
Safe-Student \cite{ossl_5} & 57.4\scriptsize{$\pm$0.5} & 60.1\scriptsize{$\pm$0.9} & 77.5\scriptsize{$\pm$0.2} & 40.9\scriptsize{$\pm$1.2} & 52.6\scriptsize{$\pm$1.3} & 61.9\scriptsize{$\pm$0.7} & 41.1\scriptsize{$\pm$1.9} & 51.4\scriptsize{$\pm$0.6} & 66.1\scriptsize{$\pm$0.3} & 77.8\scriptsize{$\pm$2.1} & 88.7\scriptsize{$\pm$0.9} \\
\textcolor{black}{OSP \cite{ossl_10}} & \textcolor{black}{68.2\scriptsize{$\pm$0.7}} & \textcolor{black}{82.3\scriptsize{$\pm$1.5}} & \textcolor{black}{83.8\scriptsize{$\pm$1.1}} & \textcolor{black}{51.4\scriptsize{$\pm$2.2}} & \textcolor{black}{58.9\scriptsize{$\pm$1.8}} & \textcolor{black}{66.3\scriptsize{$\pm$1.3}} & \textcolor{black}{52.3\scriptsize{$\pm$0.9}} & \textcolor{black}{58.8\scriptsize{$\pm$1.1}} & \textcolor{black}{67.5\scriptsize{$\pm$0.8}} & \textcolor{black}{78.1\scriptsize{$\pm$2.0}} & \textcolor{black}{85.0\scriptsize{$\pm$0.7}} \\
\textcolor{black}{SSB \cite{ossl_11}} & \textcolor{black}{76.3\scriptsize{$\pm$3.2}} & \textcolor{black}{77.8\scriptsize{$\pm$2.7}} & \textcolor{black}{90.6\scriptsize{$\pm$0.8}} & \textcolor{black}{43.2\scriptsize{$\pm$0.3}} & \textcolor{black}{54.1\scriptsize{$\pm$1.7}} & \textcolor{black}{63.6\scriptsize{$\pm$1.1}} & \textcolor{black}{42.4\scriptsize{$\pm$2.1}} & \textcolor{black}{53.9\scriptsize{$\pm$1.2}} & \textcolor{black}{67.9\scriptsize{$\pm$1.1}} & \textcolor{black}{54.9\scriptsize{$\pm$3.4}} & \textcolor{black}{85.2\scriptsize{$\pm$0.9}} \\
IOMatch \cite{ossl_7} & \textbf{90.1\scriptsize{$\pm$2.1}} & \underline{91.1\scriptsize{$\pm$1.3}} & \textbf{93.4\scriptsize{$\pm$0.4}} & \underline{53.5\scriptsize{$\pm$2.2}} & 59.6\scriptsize{$\pm$1.9} & \underline{67.0\scriptsize{$\pm$1.1}} & \underline{56.5\scriptsize{$\pm$1.2}} & 64.1\scriptsize{$\pm$0.7} & \underline{69.9\scriptsize{$\pm$0.6}} & \underline{83.9\scriptsize{$\pm$0.5}} & \underline{91.3\scriptsize{$\pm$0.3}} \\ 
UAGreg \cite{ossl_8} & 79.6\scriptsize{$\pm$4.3} & 85.4\scriptsize{$\pm$1.6} & 91.7\scriptsize{$\pm$1.4} & 52.1\scriptsize{$\pm$2.3} & \underline{60.1\scriptsize{$\pm$1.3}} & 66.2\scriptsize{$\pm$1.2} & 55.3\scriptsize{$\pm$0.7} & \underline{64.5\scriptsize{$\pm$0.6}} & 69.8\scriptsize{$\pm$0.7} & 82.1\scriptsize{$\pm$1.3} & 88.5\scriptsize{$\pm$0.8} \\
\midrule
Ours  & \underline{89.7\scriptsize{$\pm$2.2}} & \textbf{91.7\scriptsize{$\pm$1.1}} & \underline{93.3\scriptsize{$\pm$0.4}} & \textbf{53.8\scriptsize{$\pm$1.9}} & \textbf{60.5\scriptsize{$\pm$2.1}} & \textbf{67.7\scriptsize{$\pm$1.3}} & \textbf{57.0\scriptsize{$\pm$0.7}} & \textbf{65.4\scriptsize{$\pm$0.8}} & \textbf{71.1\scriptsize{$\pm$0.3}} & \textbf{84.4\scriptsize{$\pm$0.7}} & \textbf{92.1\scriptsize{$\pm$0.2}} \\
\bottomrule
\end{tabular}
}
\end{center}
\vspace{-0.2cm}
\end{table*}

\newcolumntype{b}{>{\color{black}}c}
\begin{table*}[t]
\caption{Open-set classification balanced accuracy (\%) on open-set test data across different class splits and labeled set sizes. \\Results are reported as the average with standard deviation over 3 runs on different random seeds.}
\begin{center}
\resizebox{1.0\textwidth}{!}
{
\begin{tabular}{lbccbccbcccc}
\toprule
Dataset & \multicolumn{3}{c}{CIFAR-10} & \multicolumn{6}{c}{CIFAR-100} & \multicolumn{2}{c}{ImageNet-30}  \\
\cmidrule(lr){1-1}\cmidrule(lr){2-4}\cmidrule(lr){5-10}\cmidrule(lr){11-12}
\# of konwn / unknown & \multicolumn{3}{c}{6 / 4} & \multicolumn{3}{c}{20 / 80} & \multicolumn{3}{c}{50 / 50} & \multicolumn{2}{c}{20 / 10} \\
\cmidrule(lr){1-1}\cmidrule(lr){2-4}\cmidrule(lr){5-7}\cmidrule(lr){8-10}\cmidrule(lr){11-12}
\# of labeled samples & 5 & 10 & 25 & 5 & 10 & 25 & 5 & 10 & 25 & 1\% & 5\% \\ \midrule
MTCF \cite{ossl_3} & 27.3\scriptsize{$\pm$4.2} & 33.9\scriptsize{$\pm$5.5} & 45.8\scriptsize{$\pm$0.7} & 12.7\scriptsize{$\pm$2.1} & 15.2\scriptsize{$\pm$3.1} & 27.0\scriptsize{$\pm$3.0} & 19.4\scriptsize{$\pm$1.0} & 21.9\scriptsize{$\pm$0.6} & 29.2\scriptsize{$\pm$0.5} & 24.7\scriptsize{$\pm$1.5} & 43.5\scriptsize{$\pm$0.8} \\
OpenMatch \cite{ossl_4} & 14.1\scriptsize{$\pm$0.7} & 14.5\scriptsize{$\pm$0.2} & 19.8\scriptsize{$\pm$3.4} & 12.8\scriptsize{$\pm$2.2} & 31.6\scriptsize{$\pm$1.9} & 46.8\scriptsize{$\pm$1.2} & 17.3\scriptsize{$\pm$1.1} & 40.9\scriptsize{$\pm$1.3} & 54.1\scriptsize{$\pm$1.1} & 12.8\scriptsize{$\pm$2.8} & 15.6\scriptsize{$\pm$1.9} \\
Safe-Student \cite{ossl_5} & 37.5\scriptsize{$\pm$0.9} & 43.6\scriptsize{$\pm$1.3} & 49.7\scriptsize{$\pm$0.7} & 28.9\scriptsize{$\pm$2.1} & 39.5\scriptsize{$\pm$1.8} & 49.1\scriptsize{$\pm$1.1} & 33.1\scriptsize{$\pm$1.2} & 41.1\scriptsize{$\pm$0.9} & 54.0\scriptsize{$\pm$1.2} & 61.4\scriptsize{$\pm$2.0} & 68.8\scriptsize{$\pm$1.5} \\
\textcolor{black}{OSP \cite{ossl_10}} & \textcolor{black}{40.9\scriptsize{$\pm$0.9}} & \textcolor{black}{49.4\scriptsize{$\pm$1.4}} & \textcolor{black}{50.3\scriptsize{$\pm$1.3}} & \textcolor{black}{46.3\scriptsize{$\pm$3.2}} & \textcolor{black}{52.1\scriptsize{$\pm$0.9}} & \textcolor{black}{\underline{61.2\scriptsize{$\pm$1.0}}} & \textcolor{black}{\underline{51.3\scriptsize{$\pm$1.7}}} & \textcolor{black}{56.1\scriptsize{$\pm$1.2}} & \textcolor{black}{\underline{63.1\scriptsize{$\pm$0.9}}} & \textcolor{black}{52.0\scriptsize{$\pm$1.4}} & \textcolor{black}{56.6\scriptsize{$\pm$1.2}} \\
\textcolor{black}{SSB \cite{ossl_11}} & \textcolor{black}{65.3\scriptsize{$\pm$1.5}} & \textcolor{black}{66.7\scriptsize{$\pm$1.6}} & \textcolor{black}{72.6\scriptsize{$\pm$1.2}} & \textcolor{black}{41.1\scriptsize{$\pm$2.3}} & \textcolor{black}{51.5\scriptsize{$\pm$1.7}} & \textcolor{black}{60.5\scriptsize{$\pm$1.6}} & \textcolor{black}{41.5\scriptsize{$\pm$0.8}} & \textcolor{black}{52.8\scriptsize{$\pm$0.9}} & \textcolor{black}{62.5\scriptsize{$\pm$1.3}} & \textcolor{black}{52.2\scriptsize{$\pm$1.8}} & \textcolor{black}{81.1\scriptsize{$\pm$0.9}} \\
IOMatch \cite{ossl_7} & \textbf{74.8\scriptsize{$\pm$2.3}} & \textbf{75.1\scriptsize{$\pm$2.1}} & \textbf{79.0\scriptsize{$\pm$0.3}} & \underline{46.6\scriptsize{$\pm$1.9}} & 51.4\scriptsize{$\pm$1.5} & 59.7\scriptsize{$\pm$0.5} & 50.4\scriptsize{$\pm$1.9} & \underline{56.6\scriptsize{$\pm$2.1}} & 61.6\scriptsize{$\pm$0.7} & \underline{70.3\scriptsize{$\pm$1.7}} & \underline{82.8\scriptsize{$\pm$0.6}} \\ 
UAGreg \cite{ossl_8} & 68.2\scriptsize{$\pm$1.3} & \underline{73.8\scriptsize{$\pm$1.8}} & \underline{77.5\scriptsize{$\pm$0.5}} & 45.9\scriptsize{$\pm$1.2} & \underline{54.7\scriptsize{$\pm$1.4}} & 60.5\scriptsize{$\pm$0.7} & 45.7\scriptsize{$\pm$0.8} & 55.8\scriptsize{$\pm$1.4} & 59.3\scriptsize{$\pm$0.9} & 61.8\scriptsize{$\pm$1.7} & 67.7\scriptsize{$\pm$1.2} \\ 
\midrule
Ours  & \underline{70.1\scriptsize{$\pm$3.1}} & 71.5\scriptsize{$\pm$3.3} & 75.1\scriptsize{$\pm$1.4} & \textbf{48.1\scriptsize{$\pm$1.8}} & \textbf{55.4\scriptsize{$\pm$2.2}} & \textbf{64.5\scriptsize{$\pm$1.8}} & \textbf{53.4\scriptsize{$\pm$1.2}} & \textbf{63.3\scriptsize{$\pm$1.3}} & \textbf{66.5\scriptsize{$\pm$0.9}} & \textbf{74.3\scriptsize{$\pm$1.4}} & \textbf{88.7\scriptsize{$\pm$0.5}} \\
\bottomrule
\end{tabular}
}
\end{center}
\vspace{-0.2cm}
\end{table*}
\noindent\textbf{Evaluation metrics.}
In this study, we address an open-set SSL setting where the test set comprises both known (inlier) and unknown (outlier) classes.
To assess performances on known classes, we utilize closed-set classification accuracy, measuring accuracy solely on the known classes present in the test set.
To ensure fair comparisons irrespective of varying convergence rates, we report the best results across all training epochs.
\textcolor{black}{
For evaluating performances on unknown classes, we employ open-set accuracy as our primary criterion, treating all unknown classes as a single novel class.
This metric effectively addresses errors related to misclassified samples that are overlooked by some metrics focusing solely on the binary property of inliers and outliers.
Notably, all the compared baselines adopt a detect-and-filter strategy. Accordingly, samples identified as outliers by individual algorithms are assigned to the $(K+1)$-th category.}
Considering the potential class imbalance in open-set test data, with a higher number of outliers compared to inliers from each known class, we adopt Balanced Accuracy (BA) \cite{ba} as open-set accuracy.
Following the previous work \cite{ossl_7}, we define open-set accuracy as:
\begin{equation}
\text{BA}=\frac{1}{K+1} \sum_{k=1}^{K+1} \text{Recall}_{k},
\end{equation}
where $\text{Recall}_{k}$ represents the recall score of the $k$-th class.
Each method is evaluated using its best checkpoint model based on closed-set performance.

\noindent\textbf{Implementation details.}
Similar to the standard SSL benchmarks \cite{ssl_4, ssl_5}, we adopt WideResNet-28 \cite{net_1} for small-scale experiments (CIFAR-10/100) and ResNet-18 \cite{net_2} for large-scale experiments (ImageNet) as the feature extractor $f_\text{enc}$.
The target head $f_\text{cls}$ and each divergent head $g_{k}$ consist of a single-layer perceptron that returns the expected logits for known classes $C$, with the number of divergent heads set to $K=10$.
The projection head $h$ is a two-layers MLP that maps the embedding vector $z_{b}$ to a low-dimensional space $d^{\prime}=128$.
For all experiments with our proposed method, we use an identical set of hyperparameters: \{$\lambda_{u}=1.0$, $\lambda_{\text{mi}}=0.5$, $\lambda_{\text{kd}}=1.5$, $\tau=0.95$, $\alpha=0.9$, $t_{e}=0.1$, $t_{\omega}=1.5$\}.
The size of memory bank is set to $M=256 \cdot \mu B$.

For CIFAR-10/100 datasets, we set the total number of training epochs to 256, with each epoch performing 1024 iterations of mini-batch training.
All models are trained using the SGD optimizer with a momentum of 0.9 and a batch size of $B=64$ with a relative size of $\mu=7$.
The learning rate is set to 0.03 and is gradually decreased using a cosine decay scheduler during the training process.
The weight decay is set to 0.0005 for CIFAR-10 and 0.001 for CIFAR-100, respectively.
For the ImageNet dataset, the total number of training epochs is set to 100, with each epoch performing 1024 iterations of mini-batch training.
Similar to the CIFAR settings, we use the SGD optimizer with a momentum of 0.9, while the batch size is $B=64$ with a relative size of $\mu=2$.
The learning rate is set to 0.03 with a weight decay of 0.0003, and it is gradually decreased using a cosine decay scheduler during the training process.
For the unsupervised objective, our approach incorporates two types of augmentations: a weak augmentation denoted as $\text{Aug}_w$ and a strong augmentation referred to as $\text{Aug}_s$.
The weak augmentation used in all experiments follows the standard crop-and-flip technique.
For the strong augmentation technique, we adopt the RandAugment method as $\text{Aug}_s$, as suggested in \cite{ssl_15}.
Note that for the all datasets, the same augmentation rules are applied except for the difference in crop size due to the variation in image scale.

\subsection{Main Results}
\noindent\textbf{CIFAR-10/100.}
To validate the effectiveness of proposed method on CIFAR datasets, we consider the uncorrelated setup, where unknown classes have no super-class relation with known ones, following the well-known open-set SSL benchmark \cite{ssl_1, ossl_4}.
For CIFAR-10, we define animal classes as known and other classes as unknown, resulting in 6 known and 4 unknown classes.
For CIFAR-100, we utilize the super-classes provided by the public dataset, which splits the 100 classes into 20 super-class sets.
To assess performance with varying numbers of outliers, we conduct experiments under two settings: 50 known (50 unknown) classes, and 20 known (80 unknown) classes.
\textcolor{black}{For both CIFAR-10 and CIFAR-100, we randomly select 5, 10 and 25 samples from the training set of each known class as the labeled data and the other samples, except for the labeled data, are designated as unlabeled data.
For convenience, we denote the CIFAR-10 tasks with 6 labeled classes, and 5, 10 or 25 labeled samples per class, as \textit{CIFAR-6-30}, \textit{CIFAR-6-60} or \textit{CIFAR-6-150}, respectively.}
These notations are similarly applied to other tasks.

For the closed-set classification accuracy, we compare the proposed method with both standard SSL and open-set SSL baselines, while for open-set classification accuracy, we compare only with open-set SSL baselines.
Specifically, this work aims to address cases where labeled data is underspecified, with IOMatch \cite{ossl_7} being a notable existing study in this area.
Therefore, we focus more on comparing our results with IOMatch.
Tables 1 and 2 describe the closed-set and open-set accuracy results, respectively, showing that the proposed DAC achieves the best performance in most settings.
Specifically, it significantly outperforms existing open-set SSL baselines in most cases, except for the strongest competitor, IOMatch.
The improvements are notable when class mismatch is severe and labels are scarce.
Compared to IOMatch, DAC shows substantial performance gains on CIFAR-100.
It consistently outperforms in closed-set accuracy and shows remarkable improvements in open-set accuracy against IOMatch.
However, for CIFAR-10, DAC performs similarly to IOMatch in closed-set accuracy but shows lower performance in open-set accuracy.
We presume that this suboptimal problem is due to the hyperparameter settings, since all hyperparameters for our method were tuned based on \textit{CIFAR-50-500}.
Adjusting these hyperparameters specifically for CIFAR-10 is expected to yield better performance for the proposed method.

\noindent\textbf{ImageNet.}
We evaluate the performance of our proposed approach on ImageNet, a more challenging and complex dataset.
Following \cite{ossl_4,ossl_7,ossl_8}, we utilize a subset called ImageNet-30 \cite{data_3}, which consists of 30 classes, for training instead of the complete ImageNet.
We designate the first 20 classes alphabetically as known classes, and the rest as unknown (10 classes).
For each known class, we randomly sample 1\% or 5\% images for labeled set (13 or 65 samples per class), with the remainder as unlabeled data.
Similar to CIFAR datsasets, we denote the two tasks as \textit{ImageNet-20-P1} and \textit{ImageNet-20-P5}, respectively.

The rightmost columns of Tables 1 and 2 describe the results for the ImageNet-30 dataset, confirming that our method, DAC, achieves state-of-the-art performance in both closed-set and open-set accuracy.
DAC consistently outperforms all baselines in closed-set accuracy, notably surpassing the strongest baseline, IOMatch, by 0.5\% on \textit{ImageNet-20-P1} and \textit{ImageNet-20-P5}.
With the respect to open-set accuracy, DAC markedly enhances performance, significantly outperforming existing open-set SSL baselines, including a 4.0\% and 5.9\% improvement over IOMatch on the two tasks, respectively.
These results demonstrate the effectiveness of our approach in addressing the challenges of open-set SSL problem.

\begin{table}[t]
\begin{center}
\caption{Ablation studies of the individual components.}
\resizebox{1.0\columnwidth}{!}
{
\begin{tabular}{cccc|cc|cc}\toprule
& & & & \multicolumn{2}{c}{CIFAR-100} & \multicolumn{2}{c}{ImageNet-30} \\
\cmidrule(lr){5-6}\cmidrule(lr){7-8}
D.H. & $L_\text{mi}$ & $L_\text{kd}$ & S.R. & Closed & Open & Closed & Open \\
\midrule
            &             &             &             & 59.8 & -    & 72.7 & -    \\ \midrule
\checkmark  &             &             &             & 60.4 & 41.5 & 74.1 & 50.1 \\
\checkmark  & \checkmark  &             &             & 61.3 & 54.9 & 75.3 & 62.8 \\
\checkmark  & \checkmark  & \checkmark  &             & 64.2 & 61.8 & 82.3 & 67.9 \\
\checkmark  & \checkmark  &             & \checkmark  & 63.9 & 56.1 & 78.7 & 56.0 \\
\checkmark  & \checkmark  & \checkmark  & \checkmark  & \textbf{65.4} & \textbf{63.3} & \textbf{84.4} & \textbf{74.3} \\
\bottomrule
\vspace{-0.2cm} \\
\multicolumn{8}{r}{D.H.: Divergent heads; S.R.: Soft rejection} \\
\end{tabular}
}
\end{center}
\vspace{-0.2cm}
\end{table}
\subsection{Ablation Studies}
To better understand the benefits of proposed DAC, we conducted ablation studies on quantitative and qualitative evaluations.
All experiments are performed on the \textit{CIFAR-50-500} and \textit{ImageNet-20-P1} datasets, respectively.

\noindent\textbf{Quantitative evaluations.}
Table 3 presents the numerical comparison of ablated models.
Our model uses FixMatch as the baseline objective for SSL, and its results are reported in the top row.
When only the divergent heads with $L_{\text{ssl}}$ is used, the improvements is slight due to the lack of disagreement among the heads.
Although applying the mutual information loss $L_{\text{mi}}$ significantly enhances open-set accuracy, it still shows minimal improvement in closed-set accuracy.
This is likely due to gradient blockage, which prevents effective enhancement of the feature extractor.
Notably, when the open-set knowledge distillation loss $L_{\text{kd}}$ is applied, our model achieves substantial progress in both metrics.
This demonstrates that directly propagating the uncertainty knowledge, estimated based on the proposed open-set disagreement, to the feature extractor can effectively enhance it.
Furthermore, using soft rejection results in remarkable improvements in closed-set accuracy, showing that the adaptive reweighting strategy for outliers can effectively suppress model corruption while significantly enhancing learning opportunities for valuable inliers.
Consequently, DAC surpasses the baseline, showing notable improvements of 5.6\% and 11.7\% in closed-set accuracy over the baseline in the \textit{CIFAR-50-500} and \textit{ImageNet-20-P1}, respectively.
Additionally, comparing with the second and bottom rows shows that the proposed tricks significantly improve open-set accuracy by 21.8\% and 24.2\% for both datasets.

\begin{figure*}[t]
\begin{center}
\resizebox{0.95\linewidth}{!}
{
    \begin{tabular}{c @{} c @{} c}
    \includegraphics[width=0.30\linewidth]{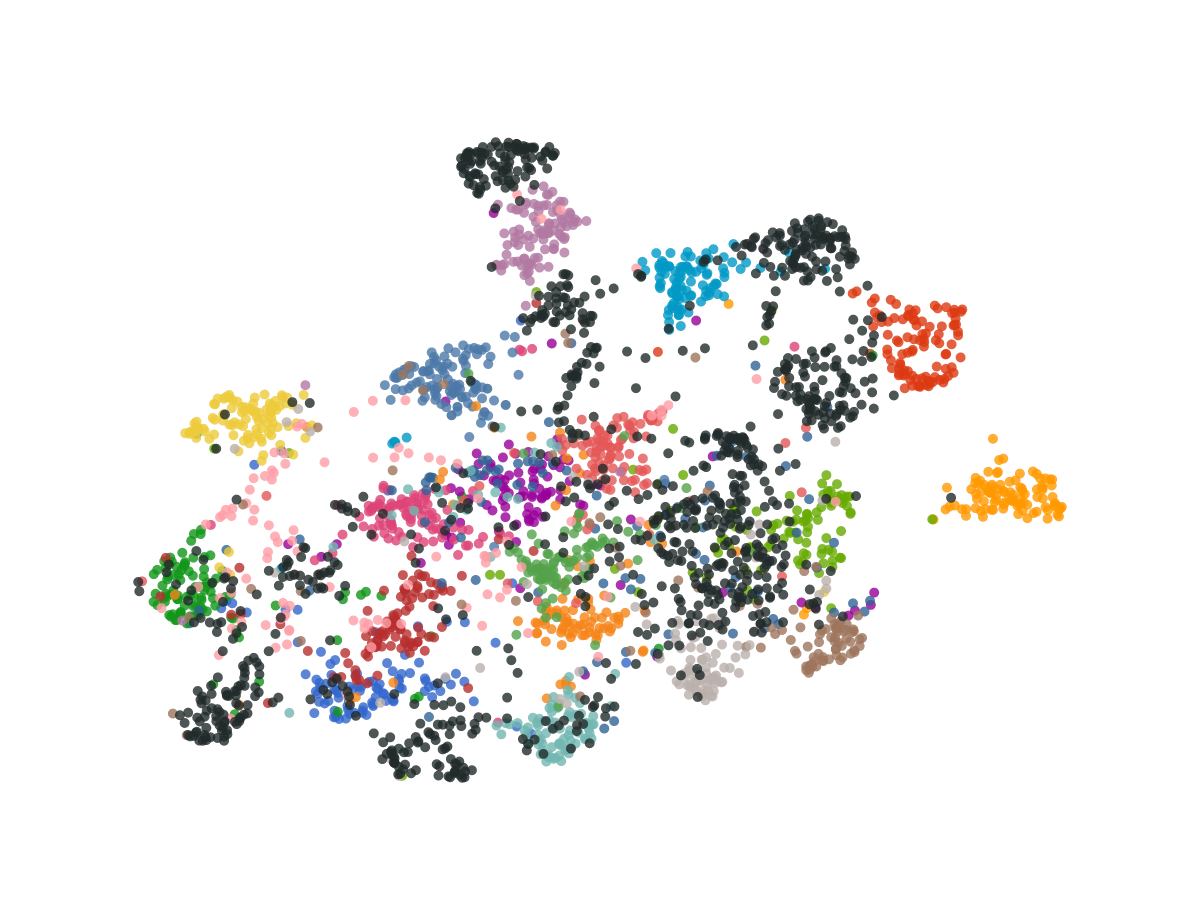} &
    \includegraphics[width=0.30\linewidth]{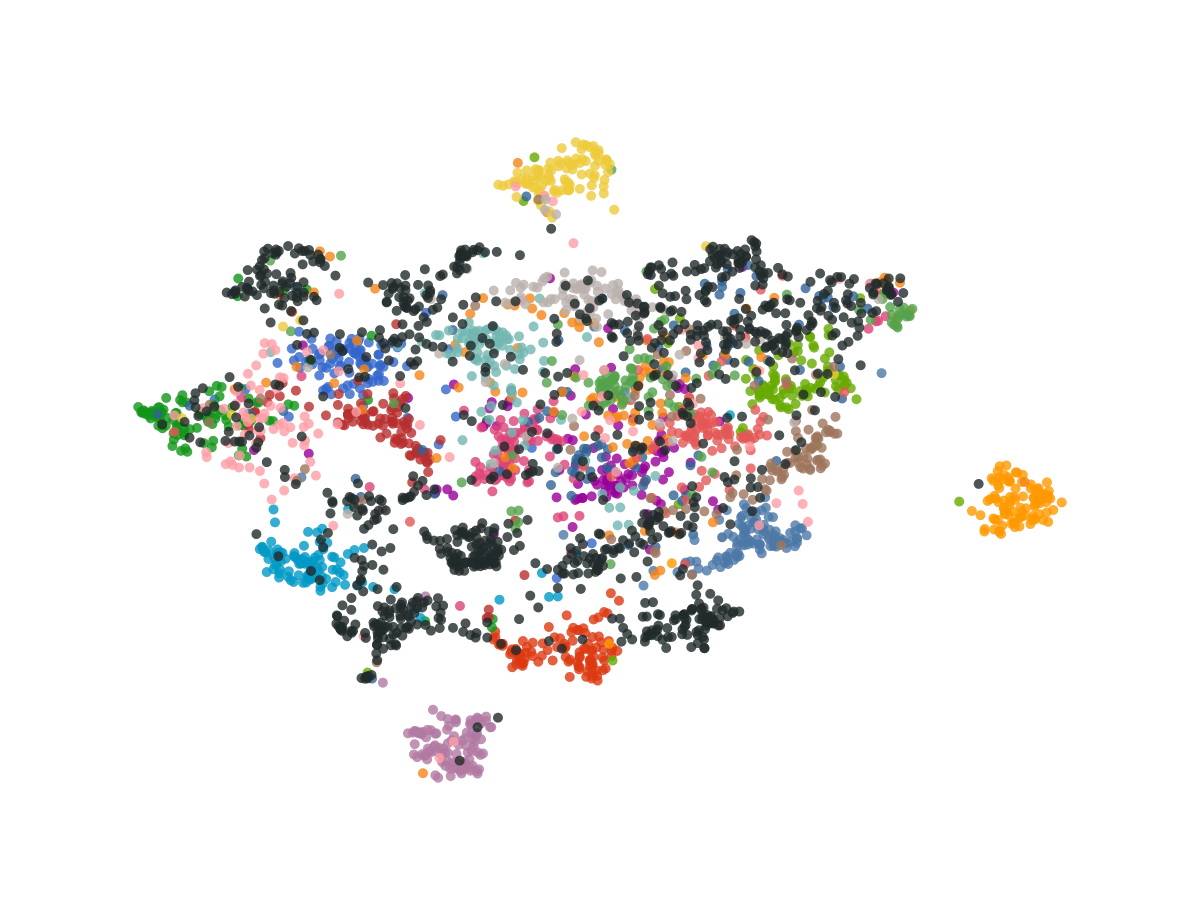} &
    \includegraphics[width=0.30\linewidth]{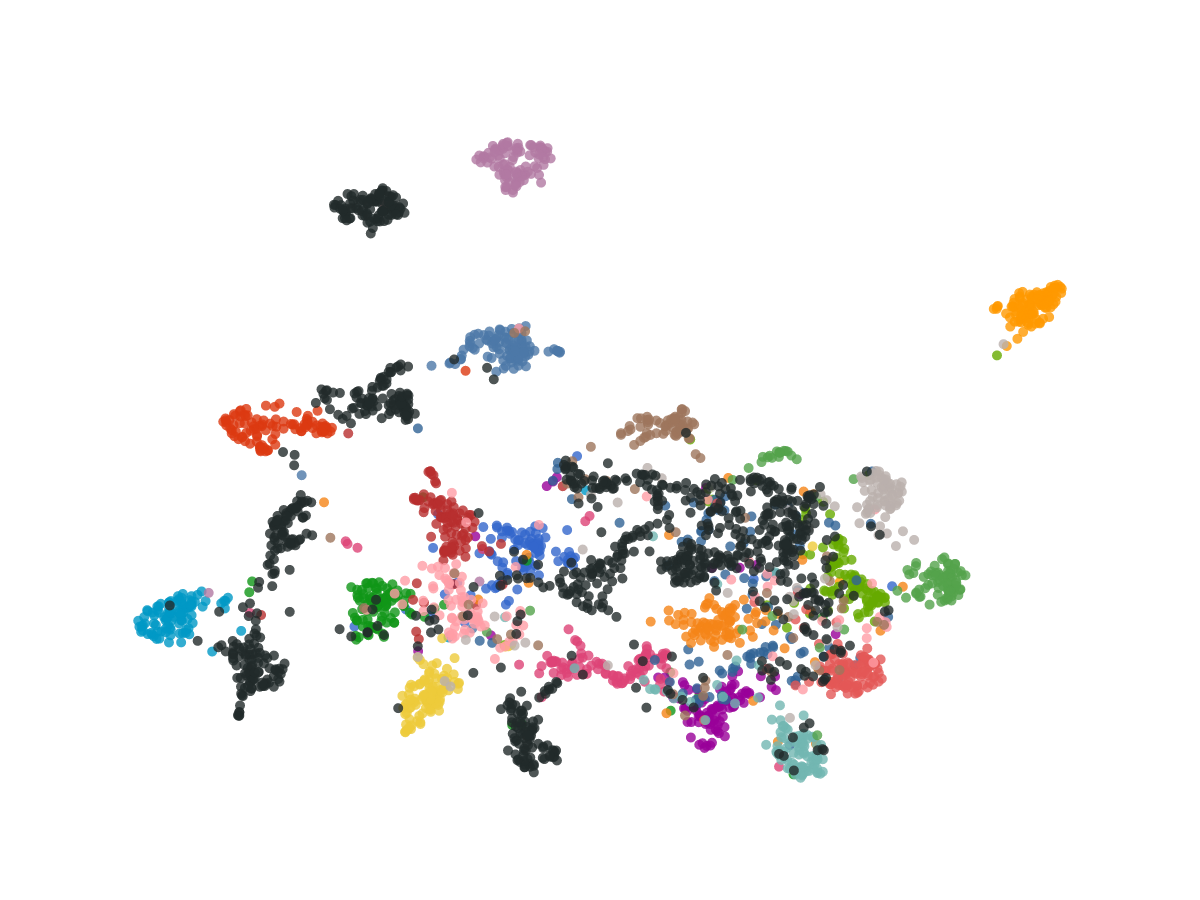} \\
    (a) w/o $L_{\text{mi}}$, $L_{\text{kd}}$ & (b) w/o $L_{\text{kd}}$ & (c) Full (DAC) \\
    \end{tabular}
}
\end{center}
\caption{UMAP visualizations of embeddings obtained from the ablated models. Black points denote outliers, while the other colored points represent distinct known classes. From left to right, the figures depict the embeddings for three different models: (a) divergent heads with $L_{\text{ssl}}$ only, (b) further applying $L_{\text{mi}}$, and (c) the final model incorporating all proposed objectives.}
\end{figure*}
\begin{figure}[t]
\begin{center}
\begin{tabular}{c @{\quad} c}
\includegraphics[width=0.45\columnwidth]
{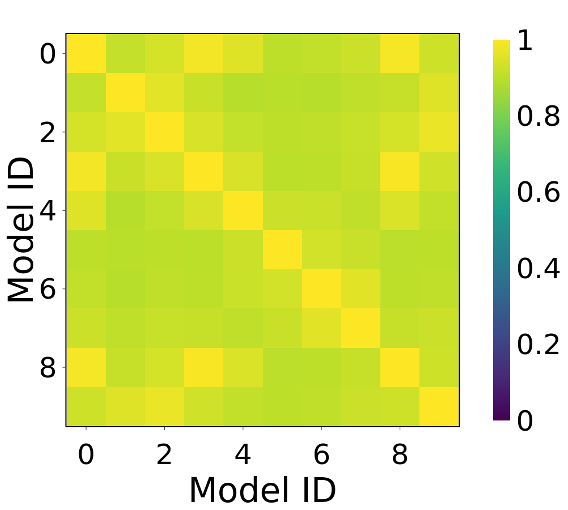} &
\includegraphics[width=0.45\columnwidth]{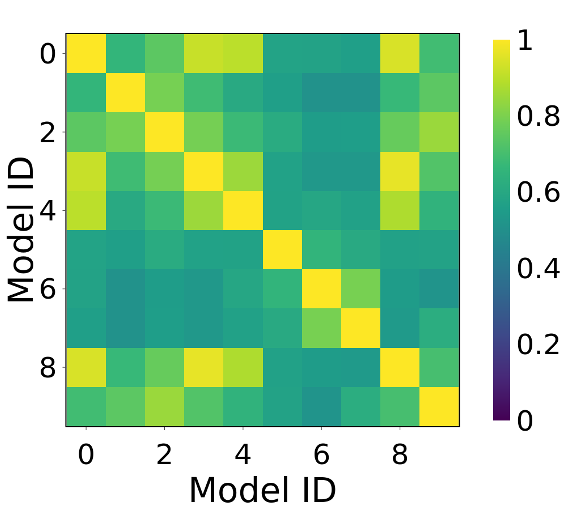}  \\
\textcolor{black}{\small{(a) Consensus for inliers}} & \textcolor{black}{\small{(b) Consensus for outliers}} \\
\end{tabular}
\end{center}
\caption{Heatmap plots of prediction disagreements on the proposed model. Each block represents the prediction consensus between a pair of heads; values close to 1 indicate that the two functions are nearly identical. Here, (a) and (b) represent the disagreement for inlier and outlier samples, respectively.}
\end{figure}
\noindent\textbf{Qualitative evaluations.}
To demonstrate the effectiveness of our proposed method, we conducted several qualitative evaluations on the test set of ImageNet-30.
First, figure 5 shows the UMAP visualizations \cite{umap} of feature embeddings $z_b$ extracted from our ablated models.
Panels (a), (b), and (c) correspond to the second, third, and fourth row models in Table 3, respectively.
Comparing (a) and (b), we observe that the proposed diversification strategy generally improves intra-class compactness for inliers, but outliers still entangle with inliers in the feature space.
In contrast, when the open-set knowledge distillation loss $L_{\text{kd}}$ is applied, we see significant improvements not only in intra-class compactness among inliers but also in the inter-class separability between inliers and outliers.
These qualitative results demonstrate that the individual components of the proposed method effectively address the open-set problem.

Then, to verify if the divergent heads form disagreements for outliers, we visualized the prediction similarity of each pair of heads as heatmaps in Figure 6.
These are consistent experiments with Figure 2 (b)-(c) in the introduction.
As shown in this figure, predictions are generally similar among head pairs for inliers but differ for outliers.
This demonstrates that the proposed diversification strategy effectively manifests the proposed perspective, open-set disagreement.
In other words, while each head is differently biased towards outliers, they maintain consistent predictions for inliers.
However, compared to Figure 2, our model shows more similar predictions overall, which we presume is due to all heads sharing the same feature extractor, unlike the independent models trained in the introduction's setting.

\begin{figure}[t]
\begin{center}
\begin{tabular}{c @{} c @{} c}
\includegraphics[width=0.33\columnwidth]{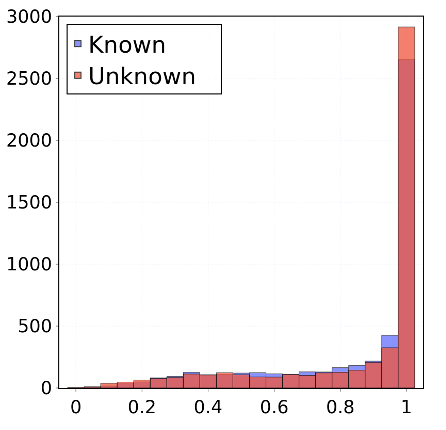} &
\includegraphics[width=0.33\columnwidth]{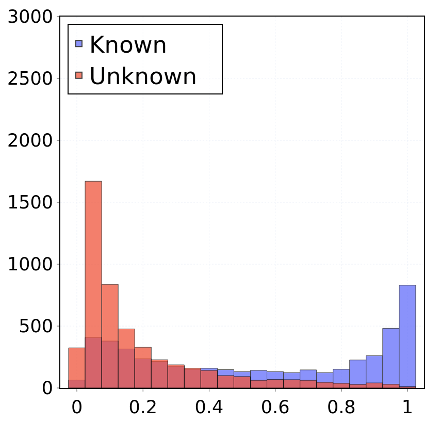} &
\includegraphics[width=0.33\columnwidth]{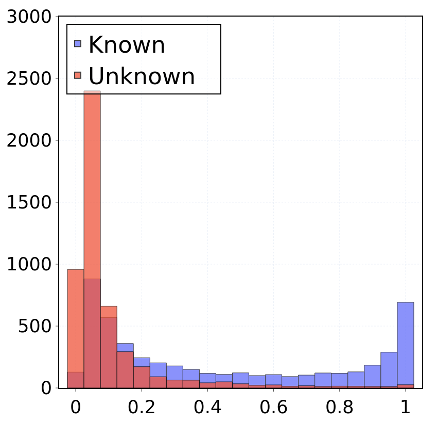} \\
(a) w/o $L_{\text{mi}}$, $L_{\text{kd}}$ & (b) w/o $L_{\text{kd}}$ & (c) Full (DAC) \\
\end{tabular}
\end{center}
\caption{Histogram plots of the consensus scores (Eq. (8)) acquired from the ablated models. Blue and red bars correspond to the inliers and outliers, respectively. From left to right, the histograms depict the uncertainty scores for three different models: (a) divergent heads with $L_{\text{ssl}}$ only, (b) further applying $L_{\text{mi}}$, and (c) the final model incorporating all proposed objectives.}
\end{figure}

Figure 7 presents the histogram visualization of the normalized consensus score from Eq. (8) estimated for each ablated model.
Figure 7 (a), (b), and (c) correspond to the second, third, and fourth row models in Table 3, respectively.
As shown in Figure 7 (a), when the proposed diversification strategy is not applied, the trained model fails to produce meaningful disagreement for outliers.
In contrast, when the proposed mutual information loss $L_{\text{mi}}$ is applied, it satisfies the open-set disagreement perspective, effectively distinguishing outliers from inliers.
This demonstrates that our diversification strategy effectively induces multiple heads to different biases.
Additionally, when the open-set knowledge distillation loss $L_{\text{kd}}$ is applied, the distinction between outliers and inliers becomes more pronounced.
This clearly supports our claims made in the previous quantitative results on Table 3.

\subsection{Sensitivity of Hyperparameters}
\textcolor{black}{To study the hyperparameter sensitivities of DAC, we conduct experiments across various parameter settings.}
All experiments are performed on the \textit{CIFAR-50-500}.

\begin{figure}[t]
\begin{center}
\resizebox{1.0\linewidth}{!}
{
\begin{tabular}{c @{\quad} c}
\includegraphics[width=0.45\columnwidth]{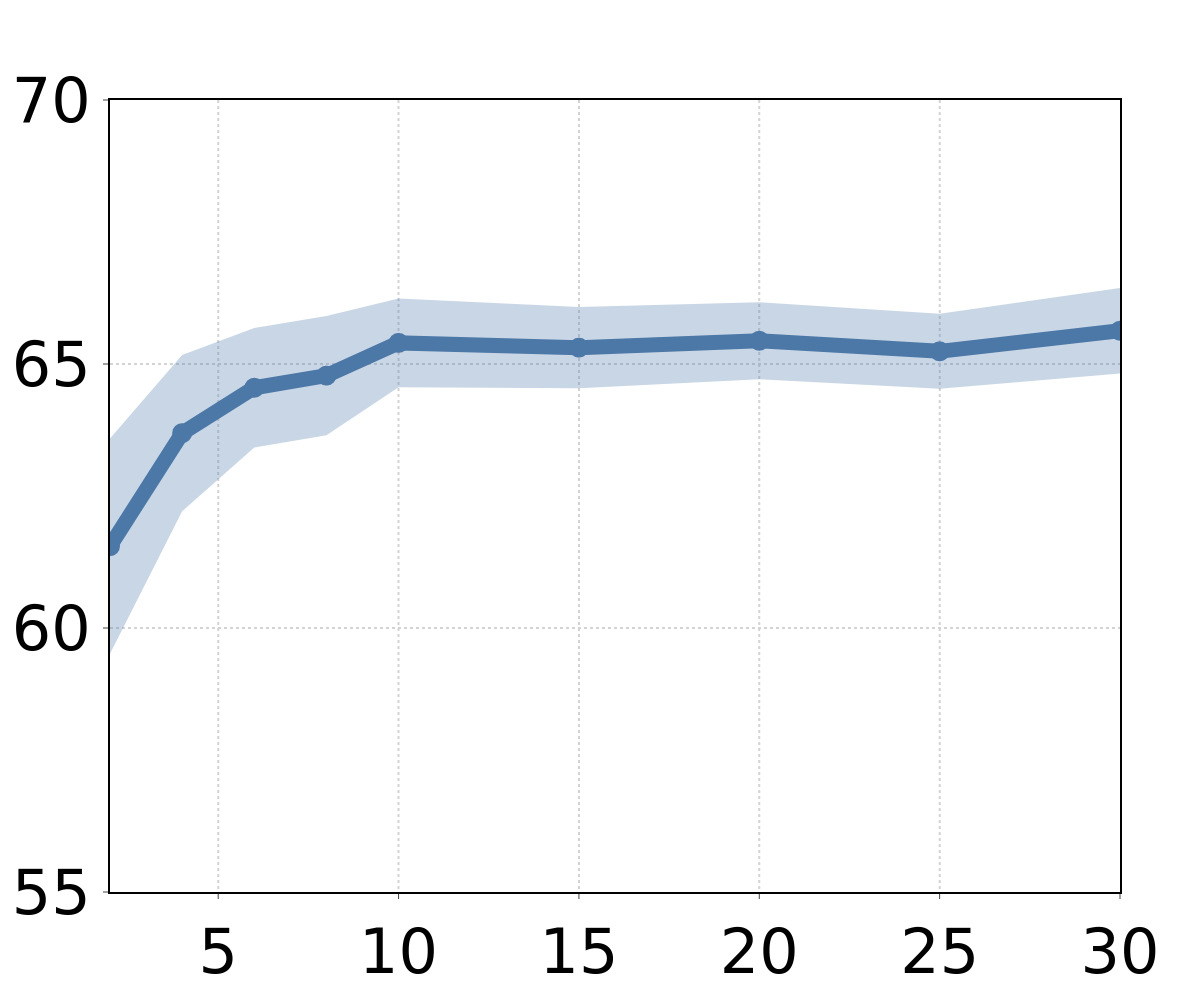} &
\includegraphics[width=0.45\columnwidth]{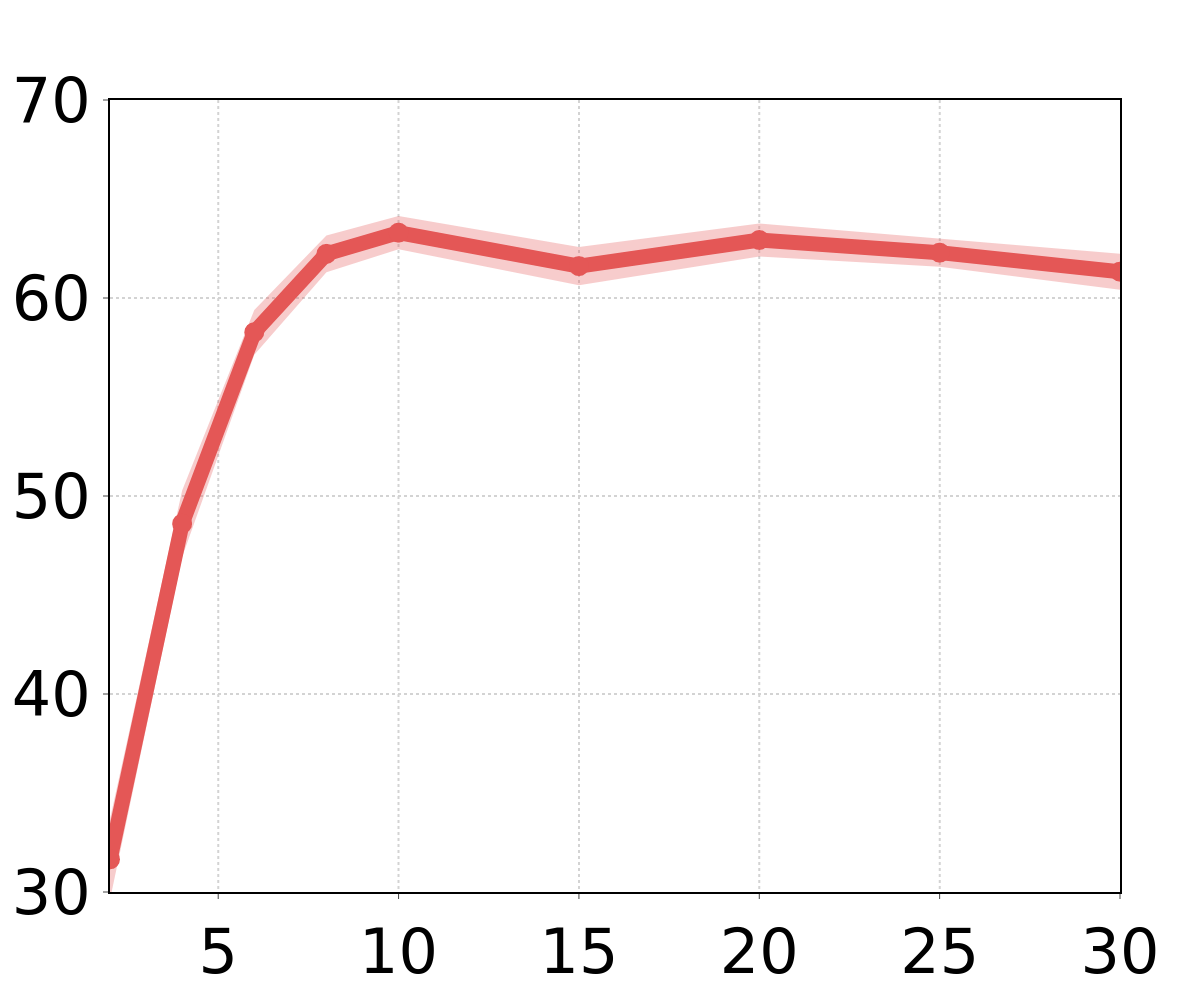}  \\
(a) Closed-set acc. & (b) Open-set acc. \\
\end{tabular}
}
\end{center}
\caption{Results of closed-set and open-set accuracy (\%) for models trained with varying numbers of divergent heads $K$ (x-axis). The experiments show that performance improves linearly up to a certain value of $K$, but beyond which the performance of model plateaus. This suggests that selecting an optimal number of heads is an efficient alternative.}
\end{figure}
\begin{figure}[t]
\begin{center}
\resizebox{1.0\linewidth}{!}
{
\begin{tabular}{c @{\quad} c}
\includegraphics[width=0.45\columnwidth]{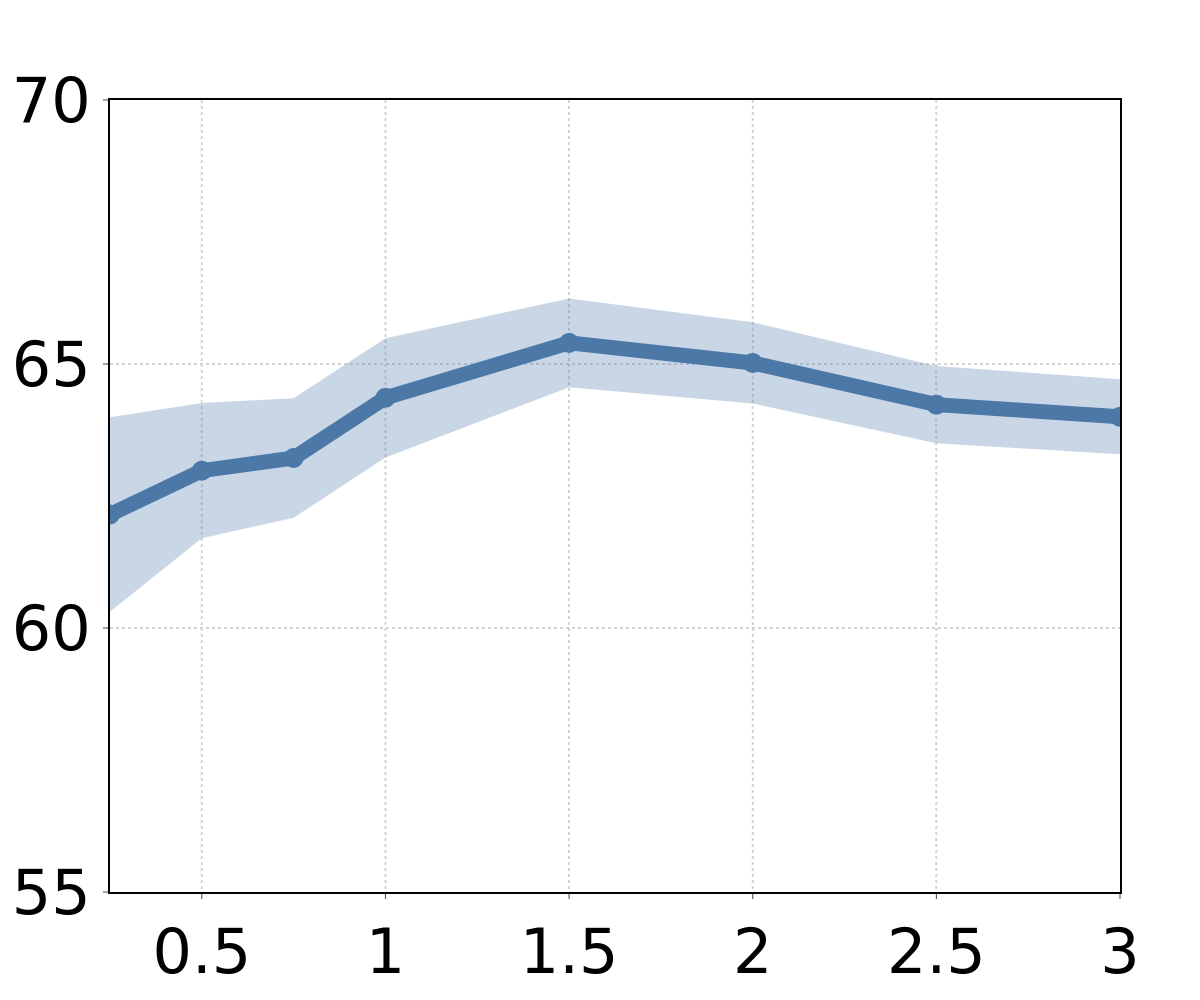} &
\includegraphics[width=0.45\columnwidth]{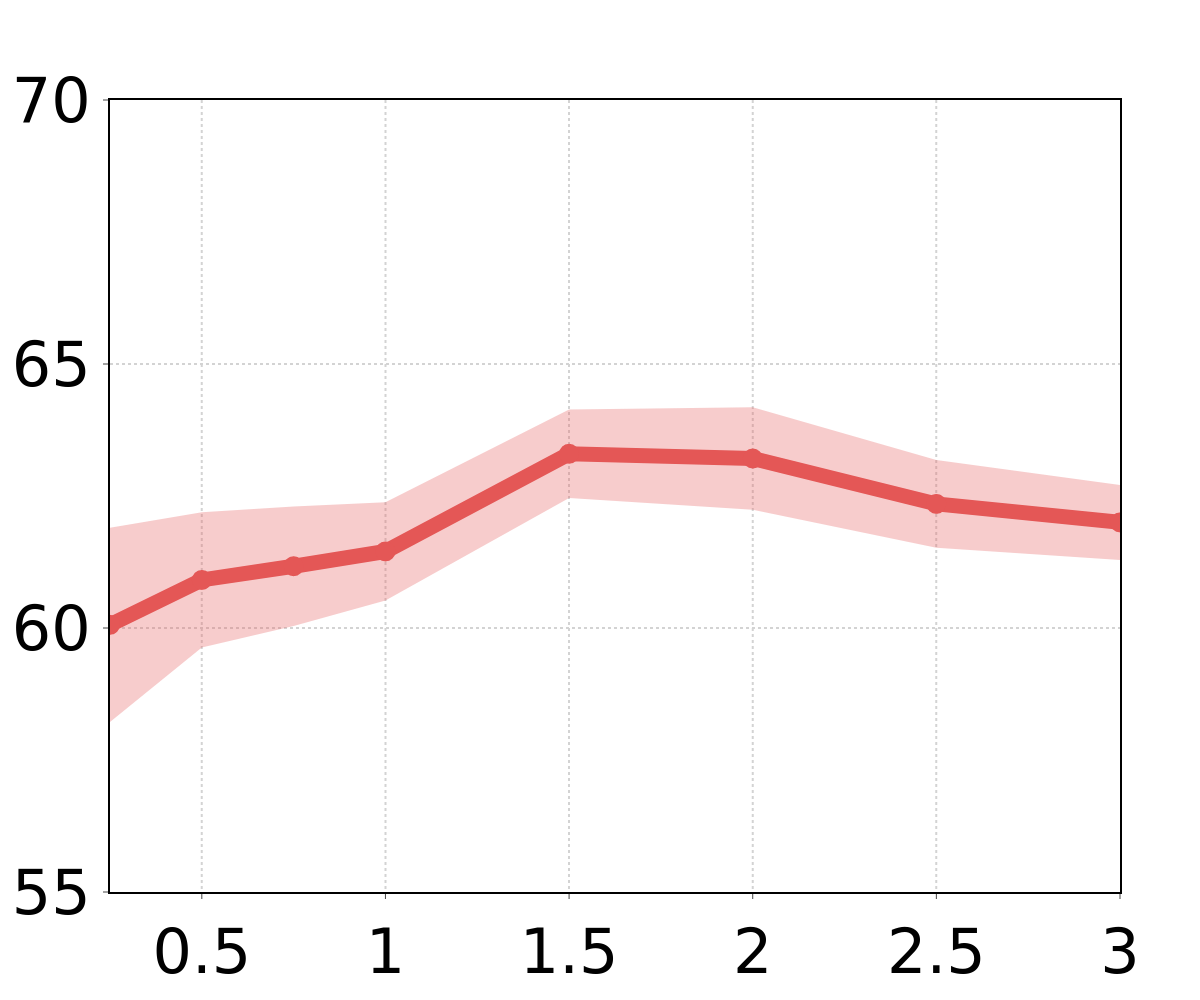}  \\
(a) Closed-set acc. & (b) Open-set acc. \\
\end{tabular}
}
\end{center}
\caption{Results of closed-set and open-set accuracy (\%) for models trained with various values of temperature $t_{w}$ (x-axis) for soft rejection sharpening. The results suggest that proposed strategy, soft rejection, is more effective for SSL performance gain compared to the hard rejection ($t_{w}=\infty$) approach.}
\end{figure}
\begin{figure}[t]
\begin{center}
\begin{tabular}{c @{} c @{} c}
\includegraphics[width=0.32\columnwidth]{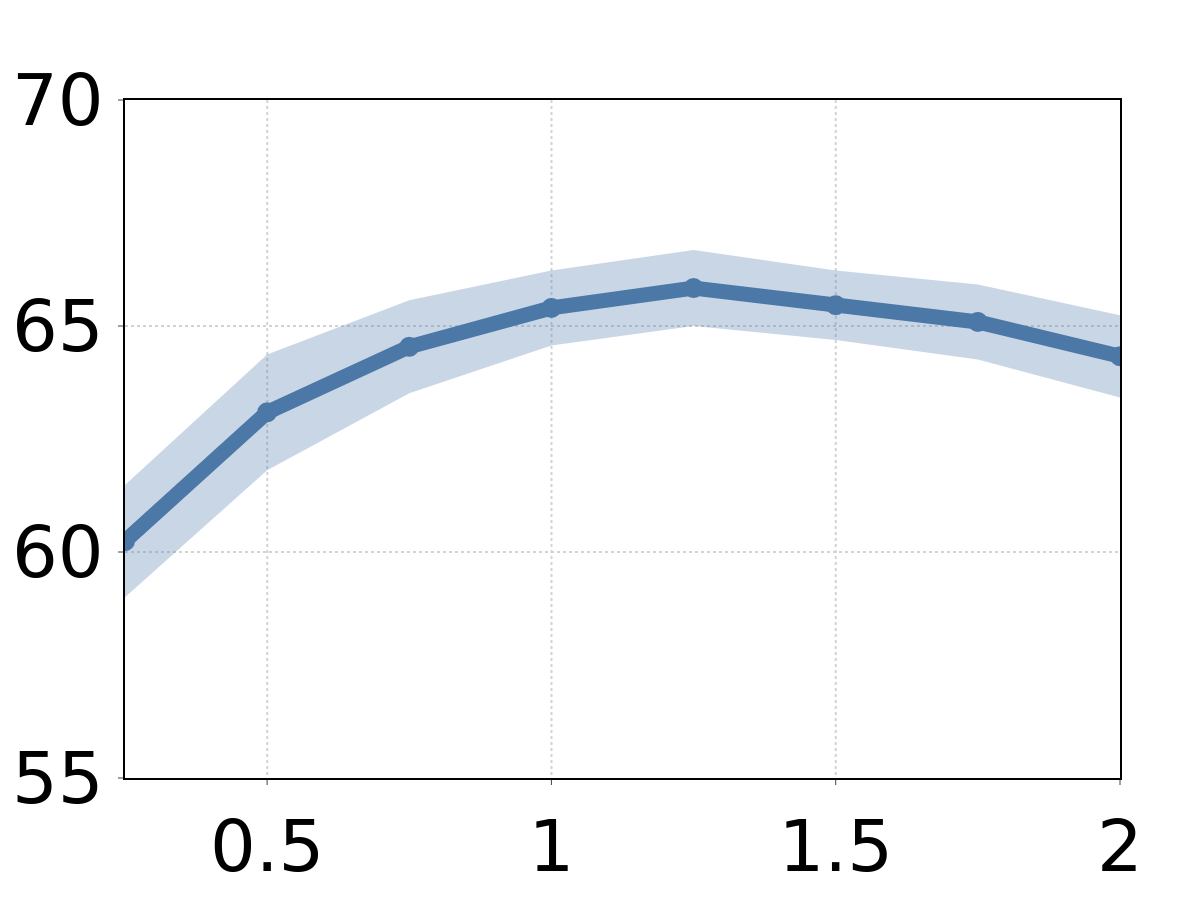} &
\includegraphics[width=0.32\columnwidth]{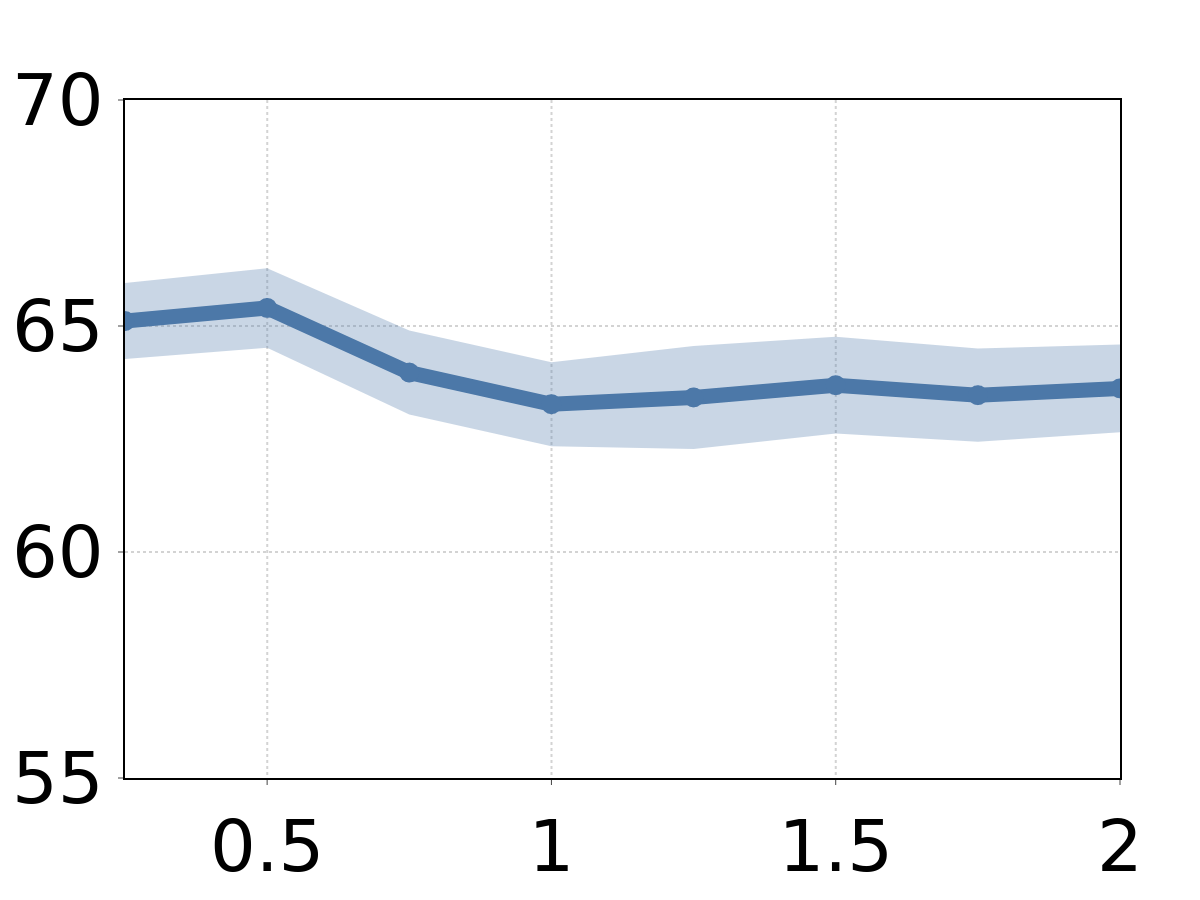} &
\includegraphics[width=0.32\columnwidth]{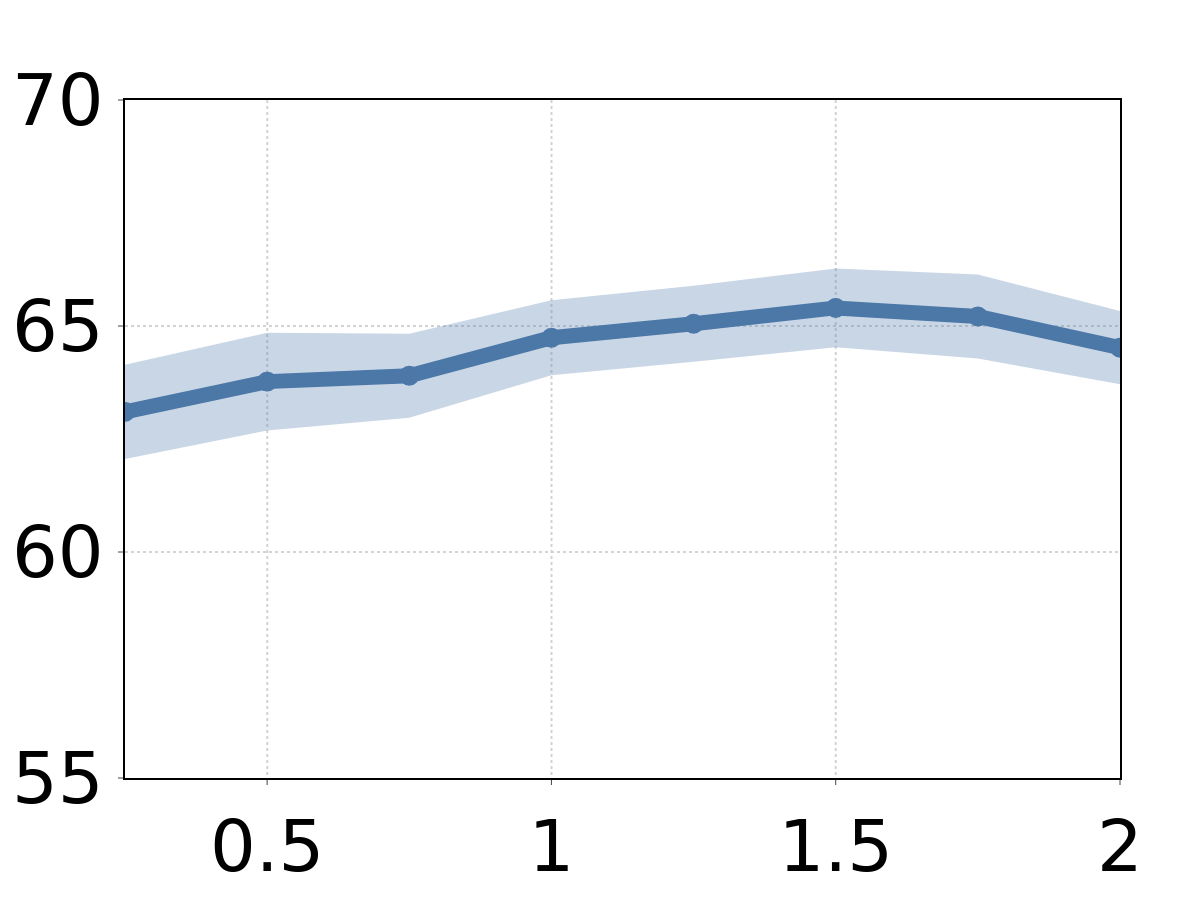} \\
(a) $\lambda_{\text{u}}$ & (b) $\lambda_{\text{mi}}$ & (c) $\lambda_{\text{kd}}$ \\
\end{tabular}
\end{center}
\caption{Analysis of hyperparameter sensitivity for the weights of objective losses ($\lambda_{u}$, $\lambda_{\text{mi}}$, and $\lambda_{\text{reg}}$). Each experiment reports the closed-set classification accuracy (\%) for different values of each weight.}
\end{figure}
\noindent\textbf{Number of divergent heads.}
Our key concept is to leverage the diversification strategy to induce the divergent heads to be differently biased, thereby forming disagreement in predictions for outliers.
Hence, we study how the number of divergent heads $K$ affects the performance of the proposed DAC.
Figure 8 shows the experimental results with various numbers of heads.
The figure demonstrates that the performance of DAC improves as the number of heads increases, but it also shows that performance tends to converge beyond a certain number of heads.
As shown in Figure 8, we attribute this result to specific head pairs converging to identical functions.
Consequently, this suggests that simply using a large number of heads cannot linearly enhance the performance of the proposed method.
Instead, finding an optimal number of heads is crucial for a more efficient strategy.
Based on these experimental results, we set the number of heads to $K=10$ for all experiments.

\noindent\textbf{Temperature scale for soft rejection.}
Figure 9 shows the experimental results for a set of different temperatures $t_{w}$, which adjust the strength of the soft rejection.
Note that as the temperature scale approaches 0, it represents standard SSL without rejecting outliers, whereas higher temperatures indicate hard rejection.
As shown in the figure, both closed-set and open-set accuracy improve with increasing temperature but start to decline slightly after reaching a maximum at a certain value.
This implies that the proposed soft rejection can generally be a better alternative than hard rejection, although finding the optimal value for this strategy is somewhat sensitive.
Based on these results, we set $t_{w}=1.5$ as the optimal value for the proposed method.

\noindent\textbf{Weights of objective losses.}
Figure 10 presents the experimental results on closed-set accuracy with varying weight parameters ($\lambda_{u}$, $\lambda_{\text{kd}}$, and $\lambda_{\text{mi}}$) used to adjust the objective losses.
For $\lambda_{u}$ and $\lambda_{\text{kd}}$, it is difficult to achieve performance improvement at values close to 0, while optimal performance is achieved at values above 1.
In contrast, for $\lambda_{\text{mi}}$, performance decreases as the value increases.
Consequently, $\lambda_{u}$ and $\lambda_{\text{kd}}$ are robust at values above 1, while $\lambda_{\text{mi}}$ is generally sensitive, necessitating careful selection.
Based on these results, we set the parameters $\lambda_{u}$, $\lambda_{\text{mi}}$, and $\lambda_{\text{kd}}$  to 1.0, 0.5, and 1.5, respectively, as optimal values.

\subsection{Further Analysis}
\begin{table}[t]
\begin{center}
\caption{Comparison results of closed-set accuracy (\%) for models integrating DAC with various SSL baselines on CIFAR-100.}
\resizebox{0.9\columnwidth}{!}
{
\begin{tabular}{lcccc}
\toprule
\# of known / unknown & \multicolumn{2}{c}{20 / 80} & \multicolumn{2}{c}{50 / 50} \\
\cmidrule(lr){1-1}\cmidrule(lr){2-3}\cmidrule(lr){4-5}
\# of labeled samples & 10 & 25 & 10 & 25 \\ \midrule
MixMatch \cite{ssl_3}  & 41.4 & 52.8 & 36.3 & 52.8 \\
w/ Ours           & 54.0 & 64.5 & 62.4 & 70.2 \\ \midrule
SimMatch \cite{ssl_6}  & 56.9 & 63.8 & 63.8 & 69.8 \\
w/ Ours           & 59.1 & 66.5 & 64.9 & 70.5 \\ \midrule
SoftMatch \cite{ssl_8} & 59.3 & 65.7 & 63.1 & 68.6 \\
w/ Ours           & 60.2 & 66.4 & 64.1 & 70.4 \\ \bottomrule
\end{tabular}
}
\end{center}
\end{table}

\noindent\textbf{Extensions on various SSL baselines.}
We demonstrate that the proposed method can be easily applied to various SSL baselines and effectively addresses the open-set problem, regardless of the SSL objective.
Therefore, we compare the performance of training models by integrating DAC into several SSL baselines.
Table 4 presents the experimental results on the CIFAR-100 dataset with 10 labeled samples per class.
As shown in the table, the proposed method consistently improves the closed-set classification accuracy across all experimental settings.
This effectively demonstrates the extensibility of the proposed method.

\begin{figure*}[t]
\begin{center}
\resizebox{0.95\linewidth}{!}
{
    \begin{tabular}{c @{\quad} c @{\quad} c}
    \includegraphics[width=0.30\linewidth]{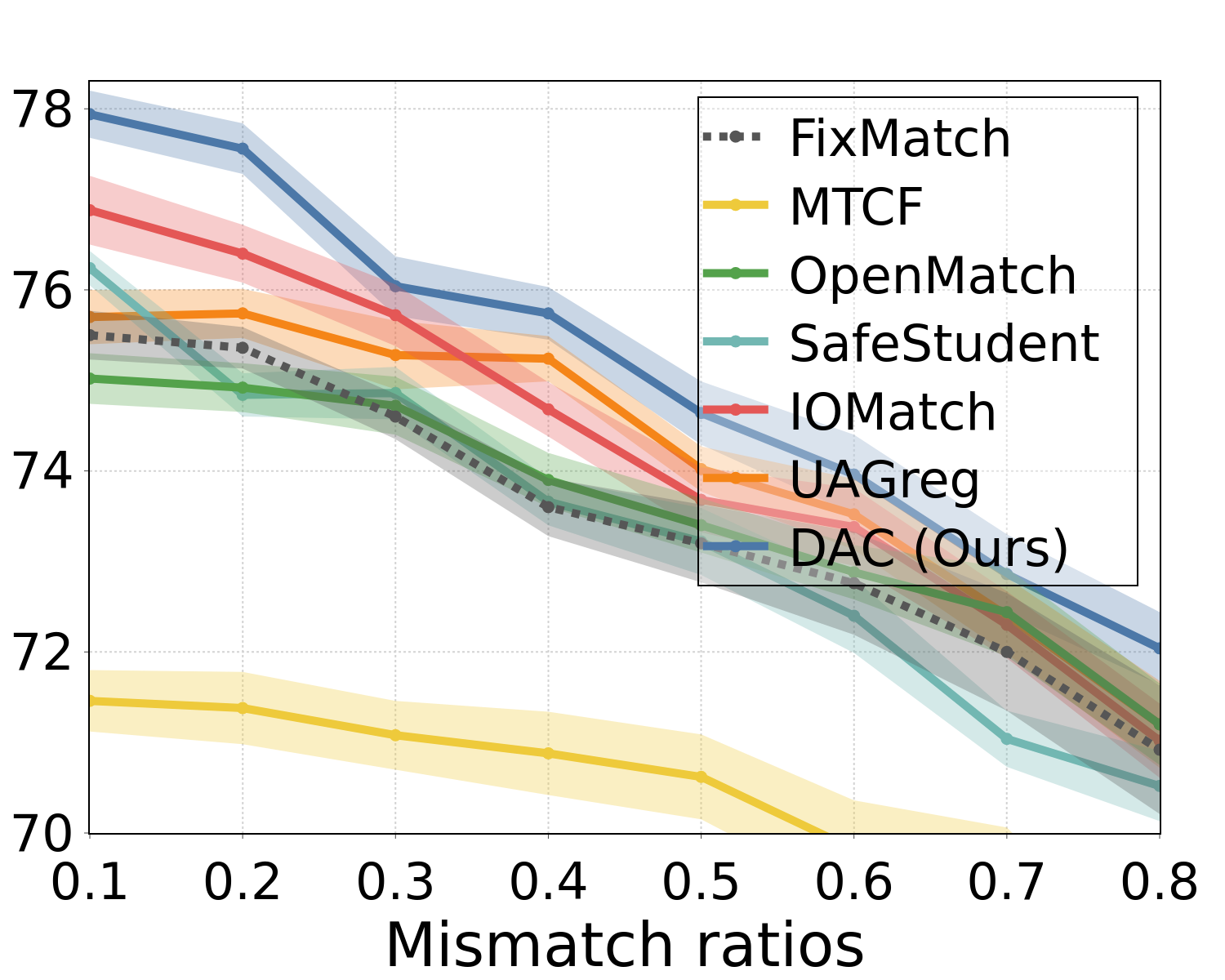} &
    \includegraphics[width=0.30\linewidth]{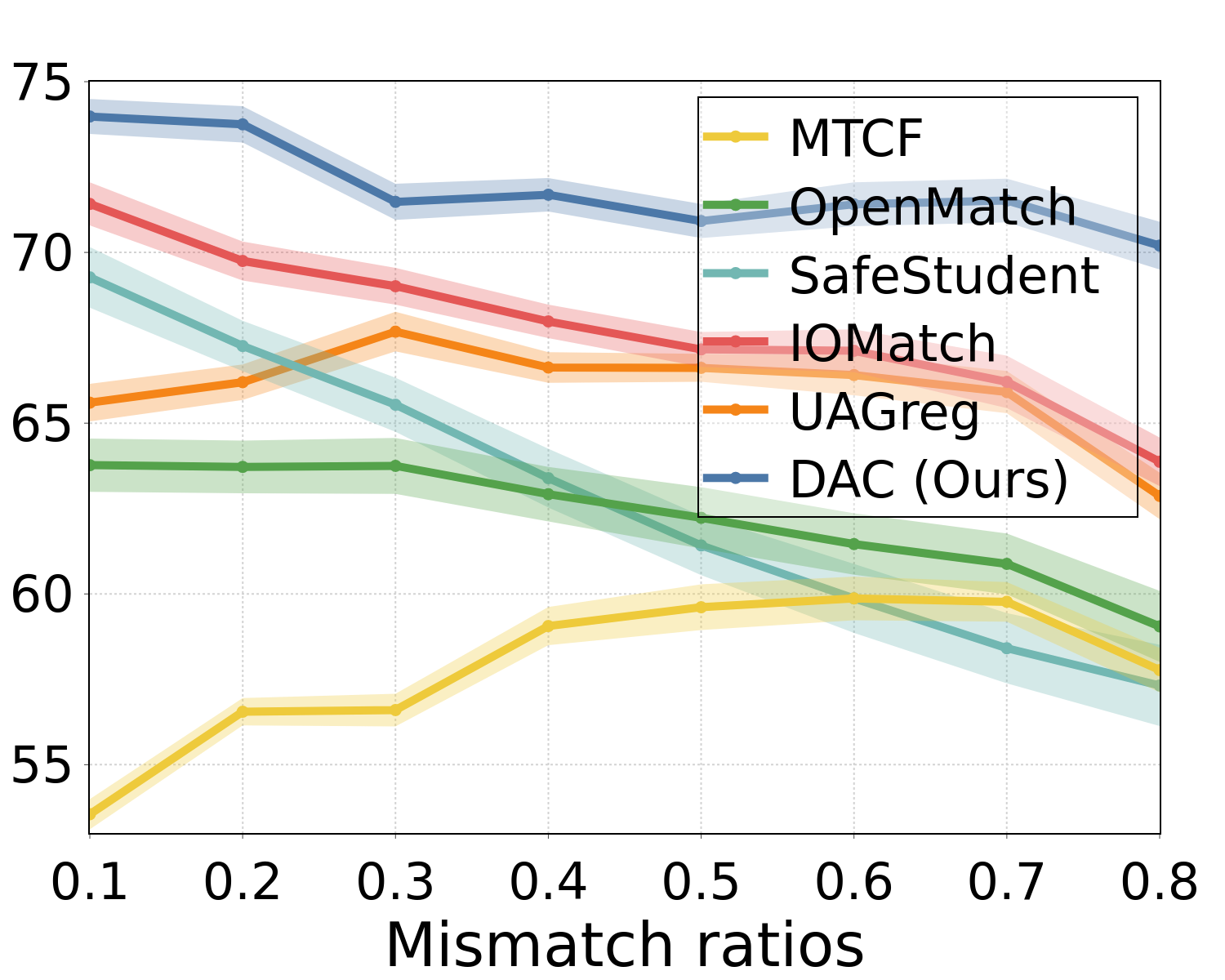} &
    \includegraphics[width=0.30\linewidth]{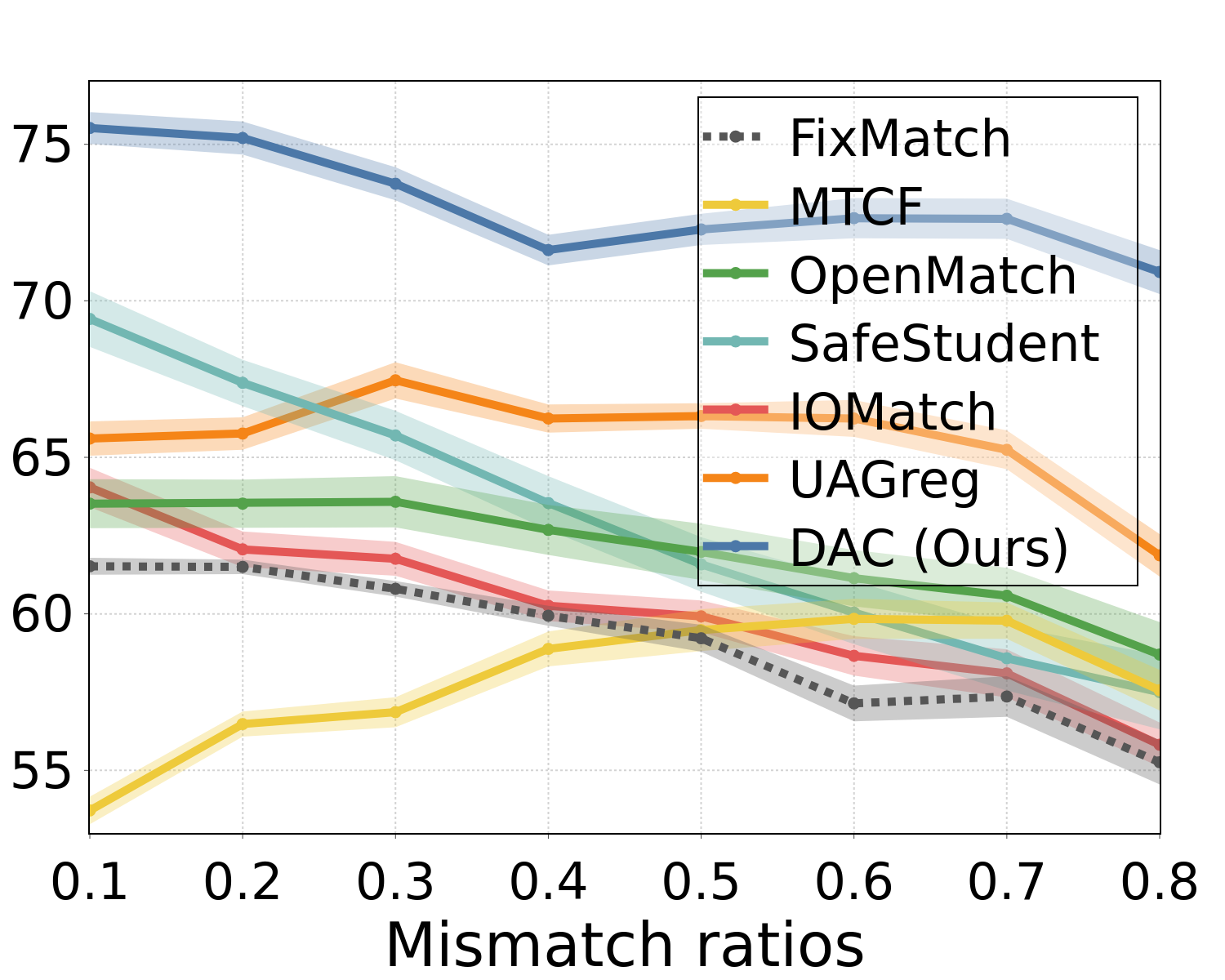} \\
    (a) Closed-set acc. & (b) Open-set acc. & (c) Recall \\
    \end{tabular}
}
\end{center}
\vspace{-0.2cm}
\caption{Comparison results for various open-set SSL algorithms trained on \textit{CIFAR-50} with varying mismatch ratios of known/unknown class distributions. The results show that the proposed DAC consistently achieves the best performance across all mismatch ratio settings. Notably, as the mismatch ratio worsens, the performance of existing open-set SSL algorithms significantly declines, while our method maintains superior performance. This suggests that our proposed strategy can effectively address underspecified cases where labeled data fails to ensure generalization over the unlabeled distribution.}
\vspace{-0.3cm}
\end{figure*}
\noindent\textbf{Results on various mismatch ratios.}
We assess the robustness of our method, DAC, against uncurated data by analyzing the performance across various mismatch ratios using the \textit{CIFAR-50} dataset with 100 labeled samples per class, similar to \cite{ossl_8}.
We consider three metrics: closed-set accuracy, open-set accuracy, and recall, as depicted in Figure 11.
The x-axis of the figure represents the mismatched ratio; for instance, a ratio of 0.3 indicates that 30\% of the unlabeled data are outliers.
Therefore, higher ratios are expected to cause greater model degradation due to uncurated data.
Across all mismatched scenarios, our proposed method consistently outperforms alternative approaches in all metrics.
The improvements are particularly notable when the mismatch ratio is severe.
Most importantly, in terms of open-set accuracy and recall, our method shows significant improvements.
The results strongly demonstrate the superiority of our proposed method in effectively addressing the open-set SSL problem, regardless of the severity of outliers.

\begin{table}[t]
\begin{center}
\caption{Comparison results of closed-set and open-set accuracy (\%) on CIFAR-100 under correlated outlier settings.}
\resizebox{1.0\columnwidth}{!}
{
\begin{tabular}{lcccc}
\toprule
\# of known / unknown & \multicolumn{2}{c}{20 / 80} & \multicolumn{2}{c}{60 / 40} \\
\cmidrule(lr){1-1}\cmidrule(lr){2-3}\cmidrule(lr){4-5}
Evaluation & Closed & Open & Closed & Open \\ \midrule
FixMatch \cite{ssl_4} & 79.5 & - & 65.0 & - \\
Safe-Student \cite{ossl_5} & 77.4 & 62.2 & 63.2 & 49.2 \\ 
IOMatch \cite{ossl_7} & \underline{83.5} & \underline{71.2} & \underline{66.9} & \underline{57.0} \\ 
UAGreg \cite{ossl_8} & 83.0 & 69.7 & 66.0 & 49.8 \\
\midrule
Ours & \textbf{83.6} & \textbf{75.7} & \textbf{68.7} & \textbf{65.7} \\
\bottomrule
\end{tabular}
}
\end{center}
\end{table}
\noindent\textbf{Results on correlated outliers.}
Following \cite{ossl_8}, we further evaluate the robustness of our DAC in correlated settings where outliers share the same super-class with inliers.
Table 5 describes the correlated CIFAR-100
closed-set and open-set results on the CIFAR-100 dataset with 10 labeled samples per class.
To assess performance under different outlier settings, we conducted experiments with 60 known (40 unknown) classes and 20 known (80 unknown) classes.
The table illustrates the consistently superior performance of our proposed method across all cases.
The results demonstrate that our method effectively addresses the open-set problem regardless of the correlation between known and unknown classes.

\begin{table}[t]
\color{black}
\begin{center}
\caption{Top-1 and Top-5 accuracy (\%) for Semi-iNat-2021.}
\resizebox{1.0\linewidth}{!}
{
\begin{tabular}{l|cc|cc}
\toprule
 & \multicolumn{2}{c}{From scratch} & \multicolumn{2}{c}{From pre-trained} \\
\cmidrule(lr){2-3}\cmidrule(lr){4-5}
Method & Top1 Acc. & Top5 Acc. & Top1 Acc. & Top5 Acc. \\ \midrule
Labeled Only      & 12.10 & 25.90 & 16.15 & 32.39 \\ \midrule
FixMatch \cite{ssl_4}  & 13.86 & 28.21 & 20.80 & 38.32 \\
CoMatch \cite{ssl_5}   & 16.16 & 31.19 & 21.62 & 38.72 \\
SimMatch \cite{ssl_6}  & 16.96 & 33.40 & \underline{22.18} & \underline{40.34} \\ \midrule
OpenMatch \cite{ossl_4} & 12.21 & 27.93 & 16.94 & 33.13 \\ 
IOMatch \cite{ossl_7}   & 14.48 & 29.38 & 20.97 & 38.25 \\ 
UAGreg \cite{ossl_8}    & \underline{17.08} & \underline{33.56} & 21.84 & 39.25 \\ \midrule
Ours              & \textbf{17.65} & \textbf{34.05} & \textbf{22.87} & \textbf{41.12} \\
\bottomrule
\end{tabular}
}
\end{center}
\end{table}
\noindent\textbf{\textcolor{black}{Results on real-world dataset.}}
\textcolor{black}{
To evaluate our approach in a more realistic data regime, we conducted additional experiments on the Semi-iNat-2021 dataset \cite{data_7}, which is designed to expose certain real-world challenges such as class imbalance, domain mismatch, and a large number of unknown classes. The dataset consists of 9,721 labeled images covering 810 distinct known categories, and 313,248 unlabeled images. Notably, our proposed model constructs multiple disagreeing heads, and thus requires memory capacity proportional to the number of classes. Therefore, in these experiments, the number of heads $K$ is set to 6.
The predefined threshold $\tau$ is set to 0.6 according to the increased number of classes, while all other hyperparameters follow the same configuration used in our previous ImageNet experiments, utilizing ResNet-18 as the feature extractor.}
\textcolor{black}{
The results are presented in Table 6. When training from scratch, our proposed model achieves the highest Top-1 and Top-5 accuracy, outperforming the strongest baseline (UAGreg) by 0.57\% and 0.49\%, respectively. Additionally, we also consider the scenario where the model is trained using self-supervised pre-training. For obtaining the pre-trained model, we exploit SimSiam \cite{rep_7} algorithm. In this case, unlike the model trained from scratch, conventional SSL algorithms surpass existing open-set SSL methods. Nevertheless, our proposed method still achieves the best performance, surpassing UAGreg by 1.03\% and 1.87\% in Top-1 and Top-5 accuracy, respectively.
These findings demonstrate that our proposed method can effectively serve as an alternative solution for addressing open-set problems, even on large-scale real-world datasets.
}

\noindent\textbf{\textcolor{black}{Limitations and future works.}}
\textcolor{black}{
Despite the effectiveness of the proposed DAC framework, several limitations remain, offering opportunities for further improvement:
\textit{i) Sensitivity to hyperparameters.} The DAC framework is sensitive to hyperparameters, particularly in the mutual information loss term. While mutual information encourages diverse predictions, it does so indiscriminately for both inliers and outliers, potentially causing excessive separation. This issue becomes more pronounced with fewer classes, as seen in datasets like CIFAR-10, where the diversity constraint in low-dimensional spaces can destabilize the model. Future work could explore adaptive or class-specific weighting strategies for mutual information and dynamic hyperparameter tuning to enhance robustness.
\textit{ii) Memory cost of the final classifier.}
While the shared feature extractor efficiently supports ensemble construction, the final multi-layer perceptron (MLP) classifier still incurs significant memory costs. This is especially evident in datasets like Semi-iNat-2021, where the classifier requires parameters proportional to the number of classes. Even with a single-layer perceptron for each head, the cost scales with the number of ensemble heads, which can become prohibitive for large datasets. Future research could explore parameter-efficient designs, such as low-rank factorization or pruning, to reduce this burden. Additionally, shared representations across heads or lightweight approximations could further alleviate the memory demands.
By addressing these challenges, the DAC framework can be further refined to improve its applicability and effectiveness across diverse real-world scenarios and machine learning tasks.
}
\section{Related Works}
\subsection{Semi-supervised Learning}
\textcolor{black}{
As a remedy for the reliance of deep supervised learning on large-scale annotated datasets, semi-supervised learning (SSL) provides effective solutions to leverage abundant unlabeled data while requiring only a small proportion of labeled samples.
Primary SSL methods can be broadly categorized into entropy minimization \cite{ssl_2, ssl_9, ssl_10}, consistency regularization \cite{ssl_11, ssl_12, ssl_13, ssl_14, tnnls_2}, and holistic approaches \cite{ssl_3, ssl_4, ssl_16}.
Among the various frameworks, FixMatch \cite{ssl_4} has garnered widespread attention as a strong baseline, offering remarkable effectiveness despite its simple training procedure.
Several studies have focused on the confidence thresholding employed by FixMatch, proposing class-specific threshold adjustments based on training difficulty \cite{ssl_19, ssl_20} or adaptive rejection strategies that consider the quantity-quality trade-off of pseudo-labels, thereby progressively filtering noisy samples \cite{ssl_8}.
In addition, some efforts have integrated FixMatch with representation learning strategies.
These techniques encompass a range of graph-based \cite{ssl_6, ssl_7, ssl_17, tnnls_3} and contrastive learning \cite{ssl_5, rep_3} approaches.
By incorporating instance-level similarity relationships into their objectives, they aim to produce more refined classifiers.
}

\textcolor{black}{
Notably, the majority of leading SSL frameworks that have demonstrated significant success are grounded in self-training \cite{ssl_2, ssl_4, ssl_10}.
That is, they leverage predictions from a weak model trained on labeled data as pseudo labels for unlabeled data.
However, conventional self-training assumes that the class distributions of the labeled and unlabeled sets are identical, rendering it incapable of handling unknown samples, \textit{i.e.}, outliers, present in the unlabeled data.
This approach consequently risks treating the outliers as known categories, ultimately yielding corrupted SSL models.
}

\subsection{Open-set Semi-supervised Learning}
\textcolor{black}{
An open-set SSL problem extends the conventional SSL setting to a more practical scenario by assuming the presence of out-of-category samples within uncurated unlabeled data.
This problem was first explored in \cite{ssl_1}, which demonstrated that existing SSL methods suffer performance degradation when outliers are present in the unlabeled set.
As an explicit solution, existing open-set SSL works have adopted a detect-and-filter strategy, that is, detecting samples that are expected to be outliers and suppressing their influence in training.
The core of this strategy lies in devising a sophisticated criterion and corresponding training procedure to effectively detect potential outliers.
One intuitive approach utilizes the predictions of a standard training model as a direct measure for outlier detection without introducing additional modules.
This broad category of methods encompasses various out-of-distribution detection techniques, such as prediction confidence \cite{ossl_2, ossl_14}, sample similarity \cite{ossl_12, ossl_16}, and energy discrepancy \cite{ossl_5, ossl_6}.
By regarding samples with uncertain predictions under the known distribution as potential outliers, these methods adopt adaptive training strategies accordingly.
Another line of the strategy is to exploit learnable detectors as an additional module to handle the outliers.
Depending on how the binary property is modeled, various methods have been proposed.
A standard way treats all known classes and unknown classes each as a single generic class, effectively employing a one-vs-one binary classifier.
Related studies have introduced a curriculum-based framework \cite{ossl_3}, self-training techniques \cite{ossl_9, ossl_10}, and contrastive learning strategies \cite{ossl_8} to train this detector.
Another way employs one-vs-all binary classifiers, which independently identify a binary property for each known class, as outlier detectors—a concept first introduced in OpenMatch \cite{ossl_4}. Subsequent studies \cite{ossl_7, ossl_11} have proposed various extensions to enhance this baseline.
In contrast, some investigations opt for more implicit strategies to reduce the impact of outliers on the training process by formalizing the training as a bi-level optimization problem \cite{ossl_1} or adopting binary decomposition \cite{ossl_13}, thereby mitigating their influence.
}

\textcolor{black}{
In this work, we adopt the detect-and-filter strategy, which has emerged as a major trend for tackling open-set SSL problem.
Our primary focus is on ensuring robust open-set SSL performance even when the labeled data is scarce, resulting in underspecified prior knowledge.
As discussed, under such underspecified regimes, standard detection-and-filter frameworks—including OpenMatch—often suffer from an over-rejection problem, excluding not only outliers but also many inliers from training.
Although several studies \cite{ossl_9, ossl_10, ossl_17, ossl_8} have indirectly leveraged potential outliers for representation learning purposes, only a few \cite{ossl_7, ossl_11} have explicitly considered solutions to the over-rejection issue.
We note that despite the lack of prior knowledge, existing alternatives \cite{ossl_7, ossl_11} still rely on the prediction uncertainty of a single detector acquired on the labeled set.
In contrast, our approach leverages the relative discrepancy between multiple functions acquired on the unlabeled data as an uncertainty measure.
This perspective enables the proposed DAC to effectively identify outliers even when the labeled data is underspecified.
}
\section{Conclusion}
In this study, we introduced a novel framework, Diversify and Conquer (DAC), to address the limitations of conventional semi-supervised learning (SSL) in the presence of outliers, which are common in real-world datasets. Our approach enhances the robustness of SSL by leveraging prediction disagreements among multiple models that are differently biased towards the unlabeled data distribution. This method allows for effective outlier detection even when the labeled data is insufficient.
Key contributions of our work include:
\textit{i)} Developing a diverse set of models through a single training process, which enables the identification of outliers by measuring prediction inconsistencies.
\textit{ii)} Utilizing estimated uncertainty scores to reduce the influence of potential outliers on the SSL objective, thereby improving model performance.
Overall, DAC offers a robust solution to the open-set SSL problem, providing a reliable alternative for scenarios where labeled data is limited and the presence of outliers is inevitable

\bibliographystyle{plain}
\bibliography{tnnls_dac}

\vspace{-1.0cm}
\begin{IEEEbiography}[{\includegraphics[width=1in,height=1.25in,clip,keepaspectratio]{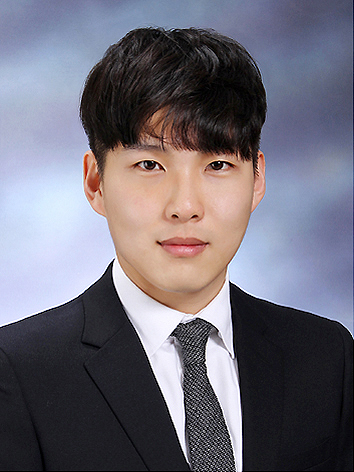}}]{Heejo Kong}
received the B.S. degree in Manufacturing Systems and Design Engineering from Seoul National University of Science and Technology, Seoul, South Korea, in 2019. He is a graduated student pursuing a Ph.D. degree in Brain \& Cognitive Engineering at Korea University, Seoul, South Korea. His research interests include artificial intelligence and learning theory of deep learning algorithm.
\vspace{-1.0cm}
\end{IEEEbiography}

\begin{IEEEbiography}
[{\includegraphics[width=1in,height=1.25in,clip,keepaspectratio]{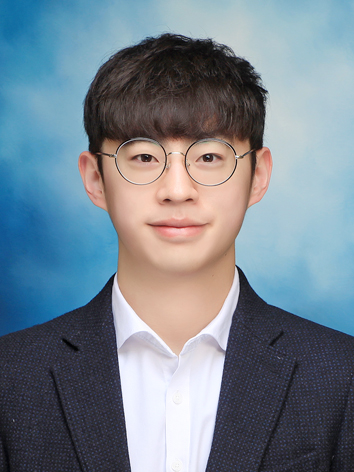}}]{Sung-Jin Kim} received his B.S. degree in mathematics and statistics from Dongguk University, Seoul, Republic of Korea, in 2021. He is a Ph.D. student in the Department of Artificial Intelligence at Korea University, Seoul, Republic of Korea. His research interests include signal processing, deep learning, and brain—computer interface.
\vspace{-1.0cm}
\end{IEEEbiography}

\begin{IEEEbiography}
[{\includegraphics[width=1in,height=1.25in,clip,keepaspectratio]{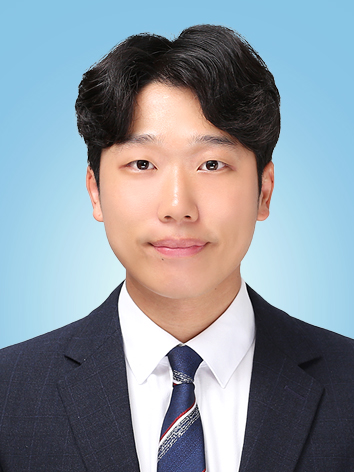}}]{Gunho Jung}
received the B.S. degree in financial mathematics from Gachon University, Seongnam, South Korea, in 2020. He is currently working toward the master’s and Ph.D. degrees with the Department of Artificial Intelligence, Korea University, Seoul, South Korea. His research interests include artificial intelligence and computer vision.
\vspace{-1.0cm}
\end{IEEEbiography}

\begin{IEEEbiography}[{\includegraphics[width=1in,height=1.25in,clip,keepaspectratio]{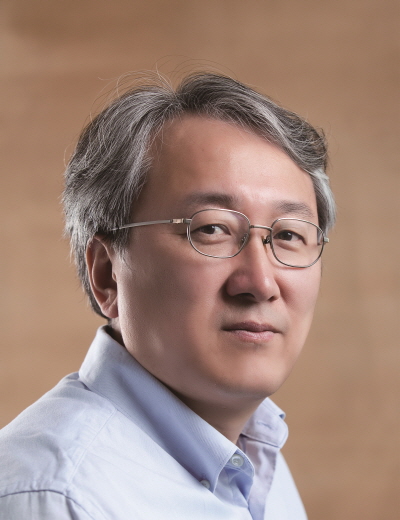}}]{Seong-Whan Lee}
(Fellow, IEEE) received the B.S. degree in computer science and statistics from Seoul National University, Seoul, Republic of Korea, in 1984, and the M.S. and Ph.D. degrees in computer science from the Korea Advanced Institute of Science and Technology (KAIST), Seoul, Republic of Korea, in 1986 and 1989, respectively. He is currently the head of the Department of Artificial Intelligence, Korea University, Seoul, Republic of Korea. His current research interests include artificial intelligence, pattern recognition, and brain engineering. Dr. Lee is a fellow of the International Association of Pattern Recognition and the Korea Academy of Science and Technology.
\end{IEEEbiography}

\end{document}